\documentclass[10pt,twocolumn,letterpaper]{article}

\usepackage[pagenumbers]{cvpr} 

\usepackage{graphicx}
\usepackage{multirow}
\usepackage{booktabs}
\usepackage{float}
\usepackage{amssymb}
\usepackage{pifont}
\usepackage{csquotes}
\usepackage[accsupp]{axessibility}
\usepackage{algorithm}
\usepackage{algpseudocode}
\usepackage{lipsum}

\usepackage{xcolor}
\definecolor{annot_green}{HTML}{0a8b4d}
\definecolor{annot_blue}{HTML}{5271FF}
\definecolor{annot_orange}{HTML}{cc6600}
\definecolor{annot_pink}{HTML}{b82edf}
\definecolor{annot_violet}{HTML}{8c52ff}

\newcommand{\xmark}{\ding{55}}%
\newcommand{\infertime}{15s }%
\newcommand{\xdownarrow}[1]{%
  {\left\downarrow\vbox to #1{}\right.\kern-\nulldelimiterspace}
}
\newcommand{\xuparrow}[1]{%
  {\left\uparrow\vbox to #1{}\right.\kern-\nulldelimiterspace}
}

\definecolor{cvprblue}{rgb}{0.21,0.49,0.74}
\definecolor{darkgreen}{HTML}{0a8b4d}
\usepackage[pagebackref,breaklinks,colorlinks,allcolors=cvprblue]{hyperref}

\def\paperID{7562} 
\def\confName{CVPR}
\def\confYear{2025}

\title{MARVEL-40M+: Multi-Level Visual Elaboration for High-Fidelity \\ Text-to-3D Content Creation}

\author{
\begin{tabular}{c c c c}
\href{mailto:sankalp.sinha@dfki.de}{\textcolor{black}{Sankalp Sinha$^{1*}$}} &
\href{mailto:mohammad.khan@dfki.de}{\textcolor{black}{Mohammad Sadil Khan$^{1*\dagger}$}} &
\href{mailto:muhammad.usama@dfki.de}{\textcolor{black}{Muhammad Usama$^{1}$}} &
\href{mailto:shino.sam@dfki.de}{\textcolor{black}{Shino Sam$^{1}$}} \\
\end{tabular} \\ [0.2cm]
\begin{tabular}{c c c}
\href{mailto:didier.stricker@dfki.de}{\textcolor{black}{Didier Stricker$^{1}$}} &
\href{mailto:skaziz.ali@hyderabad.bits-pilani.ac.in}{\textcolor{black}{Sk Aziz Ali$^{2}$}} &
\href{mailto:muhammad_zeshan.afzal@dfki.de}{\textcolor{black}{Muhammad Zeshan Afzal$^{1}$}} \\
\end{tabular} \\ [0.2cm]
\begin{tabular}{c c c c}
$^{1,2}$\text{\href{https://dfki.de/web}{\textcolor{black}{DFKI}}} & 
$^1$\text{\href{https://rptu.de/}{\textcolor{black}{RPTU Kaiserslautern-Landau}}} & 
$^1$\text{\href{https://blog.mindgarage.de/}{\textcolor{black}{MindGarage}}} & 
$^2$\text{\href{https://www.bits-pilani.ac.in/hyderabad/}{\textcolor{black}{BITS Pilani, Hyderabad}}} \\
\end{tabular} \\ [0.2cm]
\begin{tabular}{c c}
{\tt\small \href{mailto:sankalp.sinha@dfki.de}{\textcolor{black}{sankalp.sinha@dfki.de}}} & {\tt\small \href{mailto:mohammad.khan@dfki.de}{\textcolor{black}{mohammad.khan@dfki.de}}}
\end{tabular}
}

\begin{document}

\twocolumn[{%
\renewcommand\twocolumn[1][]{#1}%
\maketitle
\begin{center}
    \centering
    \captionsetup{type=figure}
    \includegraphics[width=1\linewidth]{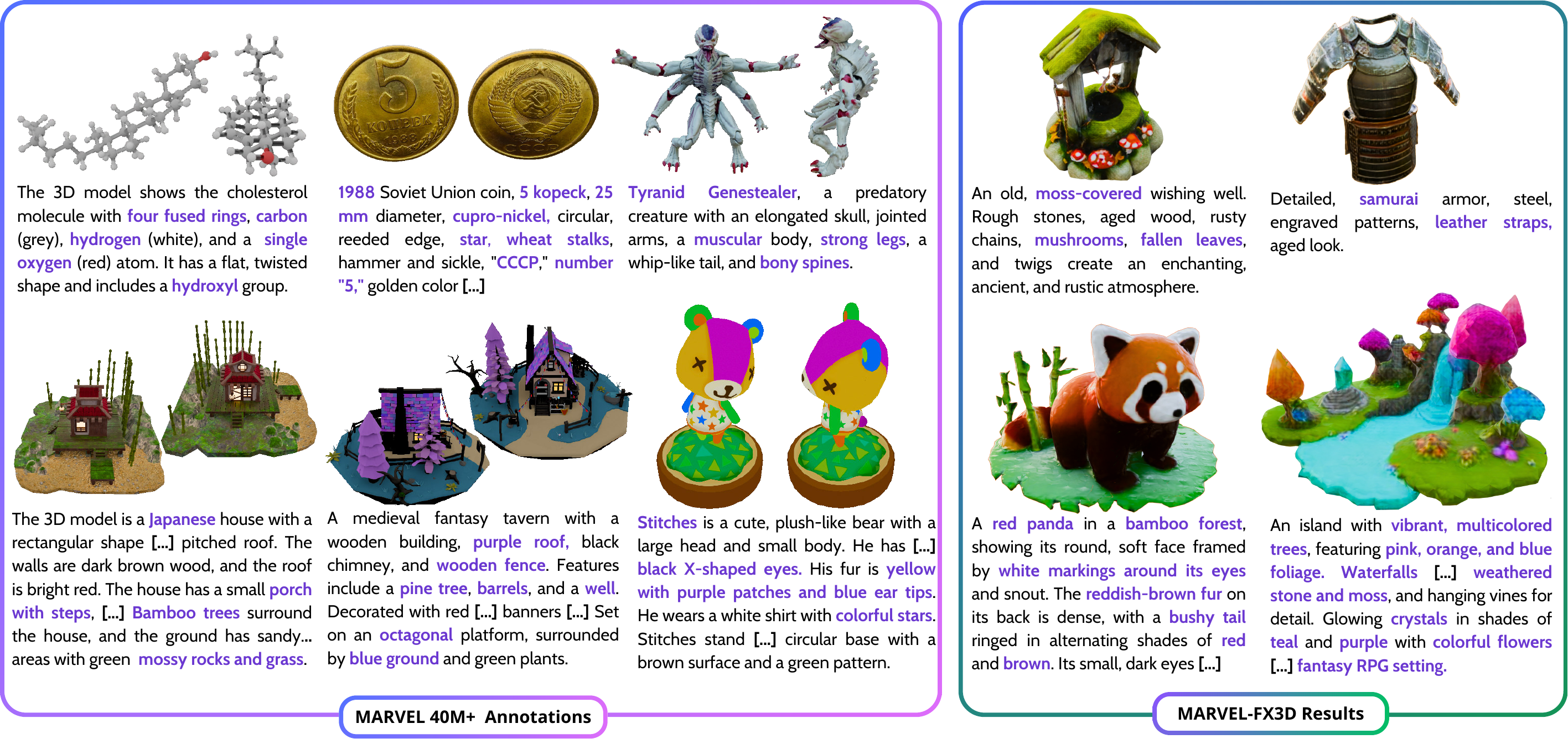}
    \captionof{figure}{\textbf{Left:} Examples of MARVEL annotations created using our proposed pipeline, which produces high-quality, domain-specific and multi-level text descriptions for 3D assets (Sec~\ref{annot_pipeline}). \textbf{Right:} Qualitative results from MARVEL-FX3D, our two-stage text-to-3D pipeline, which can generate textured mesh from text within \infertime (Sec~\ref{marvel_fx3d_architecture}). Please zoom in for details.}
   \label{fig:teaser}
\end{center}%
}]

\footnotetext[1]{Equally contributing first authors.}
\footnotetext[2]{\href{mailto:mohammad.khan@dfki.de}{\textcolor{black}{Corresponding Author.}}}

\maketitle
\begin{abstract}

Generating high-fidelity 3D content from text prompts remains a significant challenge in computer vision due to the limited size, diversity, and annotation depth of the existing datasets. To address this, we introduce MARVEL-$40$M+, an extensive dataset with $40$ million text annotations for over $8.9$ million 3D assets aggregated from seven major 3D datasets. Our contribution is a novel multi-stage annotation pipeline that integrates open-source pretrained multi-view VLMs and LLMs to automatically produce multi-level descriptions, ranging from detailed ($150$-$200$ words) to concise semantic tags ($10$-$20$ words). This structure supports both fine-grained 3D reconstruction and rapid prototyping. Furthermore, we incorporate human metadata from source datasets into our annotation pipeline to add domain-specific information in our annotation and reduce VLM hallucinations. Additionally, we develop MARVEL-FX3D, a two-stage text-to-3D pipeline. We fine-tune Stable Diffusion with our annotations and use a pretrained image-to-3D network to generate 3D textured meshes within 15s.  Extensive evaluations show that MARVEL-40M+ significantly outperforms existing datasets in annotation quality and linguistic diversity, achieving win rates of $72.41\%$ by GPT-4 and $73.40\%$ by human evaluators. Project page is available at \href{https://sankalpsinha-cmos.github.io/MARVEL/}{https://sankalpsinha-cmos.github.io/MARVEL/}.

\end{abstract} 
\vspace{-0.3cm}
\section{Introduction}\label{introduction}
\begin{table*}[t]
\resizebox{\textwidth}{!}{%
\begin{tabular}{l|lllllllll|l}
\hline
        & ShapeNet 
        & Pix3D 
        & OmniObject3D  
        & Toys4K 
        & GSO 
        & ABO 
        & Objaverse 
        & Objaverse-XL 
        & Total 3D Objects 
        & Total Captions\\ \hline
        Cap3D \cite{cap3d}   
        & \xmark & \xmark  & \xmark & \xmark & \xmark & 6,400 \cite{abo_cap3d} 
        &785,150  & 221,632 & 1,013,182 & 1,013,182 \\
3DTopia \cite{3dtopia} &     \xmark         &      \xmark        &   \xmark   &    \xmark    &  \xmark   &  \xmark   &     361,357      &      \xmark        &         361,357 &         361,357       \\
Kabra \cite{ICML2024_LeveragingVLMs} &     \xmark         &      \xmark        &   \xmark   &    \xmark    &  \xmark   &  \xmark   &     763,827      &      \xmark        &         763,827 &         763,827       \\
MARVEL  &    \textbf{52,472}      &    \textbf{374}   &      \textbf{5,878}          &    \textbf{4,000}    &  \textbf{1,030}   &  \textbf{7,953}   &     \textbf{798,759}      &       \textbf{8,031,637}       &        \textbf{8,902,103} &        \textbf{44,510,515}        \\ \hline
\end{tabular}
}
\vspace*{-.5\baselineskip}
\caption{Overview of datasets \cite{shapenet, pix3d, omniobject3d, toys4k, gso, abo, objaverse, objaversexl} annotated using our MARVEL pipeline. MARVEL provides the most extensive 3D asset annotations to date, encompassing over $8.9$M 3D objects and $40$M captions.}
\vspace*{-.5\baselineskip}
\label{tab:datasets_overview}
\vspace*{-.5\baselineskip}
\end{table*}

\noindent Text-to-3D (TT3D) content generation has emerged as a pivotal area in computer graphics, vision, and AI, enabling the creation of complex 3D objects from textual prompts \cite{li2023survey,dreamfusion,jiang2024survey} by understanding the shape, material properties \cite{assetgen,zhang2024clay}, and complex visual elaborations \cite{gala3d,wang2023prolificdreamer,Text2CAD_NeurIPS24}. This technology holds significant potential for various industries, including gaming, augmented reality (AR), virtual reality (VR), and film production \cite{li2023survey,jiang2024survey}. Recent advancements in text-to-image (TTI) synthesis \cite{sd3,sd3_huggingface,StableDiffRombach22,imagen} have achieved remarkable realism and precise control over visual effects \cite{StableDiffRombach22,ramesh2021zero,dhiman2024reflecting}. However, extending these capabilities to high-fidelity TT3D generation remains a significant challenge \cite{jiang2024survey,li2023survey,gala3d,compgs}. This is due to the intricate nature of modeling 3D shapes \cite{textmesh,magic3d,zhang2024clay,CADSigNet2024_CVPR,Text2CAD_NeurIPS24}, textures \cite{wang2023prolificdreamer,luciddreamer,magic3d}, colors \cite{zhang2024clay,assetgen} and spatial relationships \cite{gala3d,compgs} from text descriptions, a challenge further amplified by the scarcity of high-quality 3D captions. 

\vspace{0.1cm}
\noindent Current TT3D datasets like CAP3D \cite{cap3d}, 3D-Topia \cite{3dtopia}, CLAY \cite{zhang2024clay} and Kabra et al \cite{ICML2024_LeveragingVLMs} attempt to bridge this gap through automated annotations but often fall short due to their reliance on single-view VLMs \cite{llava,llava_next,blip,paliX} or GPT-4 \cite{achiam2023gpt} for caption generation. This approach frequently results in contradictory or inconsistent captions \cite{diffurank,ICML2024_LeveragingVLMs}. Moreover, the captions lack the necessary details for fine-grained 3D reconstruction. Additionally, their dependence on proprietary models like GPT-4 \cite{achiam2023gpt} introduces significant scalability and cost constraints. Manual annotation is also impractical for large-scale datasets like Objaverse \cite{objaverse} and Objaverse-XL \cite{objaversexl}. These datasets contain a diverse range of 3D models—from characters and biological elements to historical artifacts and complex ambiguous structures—requiring domain-specific expertise for accurate annotation (See Figure~\ref{fig:teaser} - Left). Beyond being time-consuming and expensive, CAP3D \cite{cap3d} has shown that human-generated captions may not necessarily surpass automated methods in quality.

\vspace{0.1cm}
\noindent To address the previously mentioned challenges, we introduce MARVEL(\textbf{M}ulti-Level \textbf{V}isual \textbf{ELA}bo\textbf{R}ation), an automated and scalable 3D captioning pipeline. Our approach combines state-of-the-art multi-view VLM InternVL2 \cite{internvl,internvl1.5} and Qwen $2.5$ LLM \cite{qwen2} to generate high-quality captions for over $8.9$ million 3D models across seven datasets \cite{shapenet,pix3d,omniobject3d,toys4k,gso,abo,objaverse,objaversexl}. To ensure domain specific information into our captions and reduce VLM hallucinations, we integrate human metadata from source datasets into our pipeline. Following \cite{chen2024single,zhuang2024gtr}, we identify five key aspects for fine-grained 3D reconstruction: object names and components, shape and geometry, texture and materials, colors, and contextual environments. Our pipeline progressively compresses these aspects to generate five levels of annotations, ranging from comprehensive descriptions ($\sim$$200$ words) for fine-grained 3D reconstruction to concise tags ($\sim$$10$ words) for quick modeling, resulting in $40+$ million annotations. Our pipeline addresses three fundamental challenges in 3D captioning - detail, accuracy, and scalability. Through comprehensive experimental analysis, we show that MARVEL-40M+ has superior annotation quality, information density, and linguistic diversity compared to other methods \cite{cap3d,ICML2024_LeveragingVLMs,3dtopia}.

\vspace{0.1cm}
\noindent  To showcase the application of our dataset, we introduce MARVEL-FX3D (\textbf{F}ast e\textbf{X}ecution for 3D), a two-stage pipeline designed for high-fidelity TT3D generation. In the first stage, we fine-tune Stable Diffusion (SD) 3.5 \cite{sd3_huggingface} with our annotations to improve its capability to produce images for suitable 3D reconstruction. In the second stage, we leverage the pre-trained Stable Fast 3D (SF3D) \cite{sf3d} for rapid image-to-3D conversion. This enables the creation of textured meshes from texts within \infertime. Our approach is inspired by multi-stage TT3D pipelines \cite{assetgen,instant3d,pointe}, a promising direction \cite{jiang2024survey,li2023survey,instant3d} that addresses the limitation of existing Score Distillation Sampling (SDS)\cite{dreamfusion}-based methods like \textit{janus problem} \cite{luciddreamer,magic3d,dreamfusion,wang2023prolificdreamer}, oversaturation \cite{luciddreamer}, and lengthy per-prompt optimization \cite{luciddreamer,magic3d,dreamfusion,wang2023prolificdreamer,hifa,textmesh}. Our experiments demonstrate that MARVEL-FX3D outperforms current state-of-the-art TT3D methods in terms of prompt fidelity and overall preference.


\vspace{0.1cm}
\noindent Our \textbf{contributions} can be summarized as follows:
\begin{enumerate}
    \item  We present MARVEL, an automated, scalable annotation pipeline for generating high-quality 3D captions. To the best of our knowledge, MARVEL-$40$M+ is the largest 3D caption dataset to date.
    \item We propose a multi-level annotation structure that spans from detailed descriptions for fine-grained 3D reconstruction to concise tags for quick modeling. 
    \item We incorporate human metadata from source datasets into our pipeline to inject domain-specific information in the text descriptions and reduce VLM hallucinations.
    \item As a downstream task, we introduce MARVEL-FX3D, a two-stage framework for high-fidelity TT3D generation.
    \item Thorough experiments demonstrate that MARVEL-40M+ achieves state-of-the-art performances in linguistic diversity, image-text alignment, caption accuracy, and high-fidelity TT3D generation. 
\end{enumerate}


\vspace*{-.5\baselineskip}
\section{Related Work} \label{related_work}

\noindent \textbf{3D-Text Data:}
3D datasets such as ShapeNet \cite{shapenet}, Objaverse~\cite{objaversexl,objaverse}, and Omniobject3D~\cite{omniobject3d} have played a crucial role in advancing 3D understanding tasks such as single~\cite{yan2016perspective,hong2023lrm,liu2024one,zero123} or multi-view~\cite{yang2024consistnet,shi2024mvdream} 3D reconstruction, multi-view consistent image generation~\cite{liu2023syncdreamer,yang2024dreamcomposer,viewdiff}, and 3D object synthesis~\cite{zhang2024clay,MakeShapePMLR24}. However, they often lack meaningful language descriptions, with available metadata being either noisy or inadequate~\cite{cap3d,openshape}. This language-3D gap has been a major bottleneck in developing high-fidelity TT3D models~\cite{cap3d,3dtopia,ICML2024_LeveragingVLMs}. Recent works like CAP3D~\cite{cap3d} addresses this by proposing an automated pipeline. It starts with BLIP~\cite{blip2} for single-view captioning of 3D assets followed by refinement using CLIP~\cite{clip} and caption aggregation by GPT-4~\cite{achiam2023gpt}. Subsequent works, Kabra et al.\cite{ICML2024_LeveragingVLMs} introduced ScoreAgg and PaLI-X\cite{paliX} to improve caption accuracy, while 3D-Topia~\cite{3dtopia} explored an alternative path with LLaVA~\cite{llava,llava_next} and GPT-3.5. CLAY~\cite{zhang2024clay} took a more direct approach, leveraging GPT-4~\cite{achiam2023gpt} for multi-view caption generation. Yet, all these approaches face inherent trade-offs. Single-view VLM approaches~\cite{cap3d,ICML2024_LeveragingVLMs,3dtopia,diffurank} often produce incomplete or inaccurate annotations~\cite{ICML2024_LeveragingVLMs,diffurank} for 3D models, while GPT-4-based methods~\cite{zhang2024clay,achiam2023gpt,cap3d,3dtopia} struggle with scalability and cost\cite{gpt4_no_scale}. Our work presents a novel solution to these challenges through three key innovations. First, we leverage open-source multi-view VLM InternVL2~\cite{internvl,internvl1.5} and Qwen $2.5$ LLM~\cite{qwen2}, achieving GPT-4~\cite{achiam2023gpt} comparable performance~\cite{internvl,internvl1.5,qwen2,open_vlm_leaderboard} without its scalability and cost constraints. Second, instead of discarding human metadata from source datasets as done in previous works~\cite{cap3d,3dtopia,ICML2024_LeveragingVLMs,zhang2024clay}, we recognize its value as domain-specific prior knowledge. We incorporate filtered versions of this metadata into our pipeline to inject relevant context and reduce VLM hallucinations. Finally, we introduce a hierarchical annotation framework with five distinct levels, ranging from detailed descriptions to abstract tags. This multi-level approach represents a significant departure from existing methods~\cite{cap3d,3dtopia,ICML2024_LeveragingVLMs,zhang2024clay}, which typically provide only single-level annotations.

\vspace{0.1cm}
\noindent \textbf{Text-to-3D:} Current TT3D methodologies can be broadly categorized into two main approaches. One prominent direction is based on the seminal work of DreamFusion~\cite{dreamfusion}, which introduced Score Distillation Sampling (SDS) to learn a NeRF~\cite{MildenhallNeRF2020} representation by leveraging information from pretrained TTI models~\cite{StableDiffRombach22,imagen,deepfloydif}. Subsequent studies have advanced this framework by improving training stability~\cite{wang2023prolificdreamer,luciddreamer},increasing output diversity~\cite{luciddreamer,hifa,wang2023prolificdreamer} and geometry extraction~\cite{magic3d,textmesh,gsgen,gaussiandreamer}. However, SDS-based methods face two key challenges: geometric inconsistencies known as the \textit{Janus problem}~\cite{jiang2024survey,li2023survey,geom_sds} and slow optimization times. This issue has been partially addressed using amortization efforts~\cite{latte3d,att3d}. The second group of methods consists of multistage pipelines~\cite{assetgen,instant3d,pointe,shap_e,flex3d,dual3d}. The goal is to generate single or multi-view images from a TTI model~\cite{StableDiffRombach22,imagen,deepfloydif,sd3,sd3_huggingface}, followed by view reconstruction into various 3D representations~\cite{MildenhallNeRF2020,flex3d,instant3d,shi2024mvdream,sf3d,hong2023lrm}. These methods often fine-tune the TTI~\cite{StableDiffRombach22,imagen,deepfloydif,sd3,sd3_huggingface} models on TT3D datasets~\cite{cap3d,ICML2024_LeveragingVLMs} to align the output image with reconstruction needs. Point-E~\cite{pointe} fine-tunes GLIDE~\cite{glide} for TTI synthesis and uses a point diffusion transformer for 3D point cloud generation. Instant3D~\cite{instant3d} fine-tunes SD~\cite{StableDiffRombach22} to produce a $2\times2$ grid of multi-view images and uses LRM~\cite{hong2023lrm} for 3D Gaussian reconstruction. AssetGen~\cite{assetgen} extends LRM towards high-quality 3D meshes with detailed textures and PBR materials. Our dataset, MARVEL-$40$M+, is uniquely positioned to advance this domain by providing comprehensive, high-quality, and domain-specific text annotations that bridge the gap between TTI generation and image-to-3D reconstruction. By fine-tuning on MARVEL-$40$M+, we develop MARVEL-FX3D, which demonstrates better performance for high-fidelity TT3D generation compared to existing state-of-the-art methods.

\vspace*{-.6\baselineskip}
\section{Methodology} 
\begin{figure*}[t]
    \centering
    \includegraphics[width=1\linewidth]{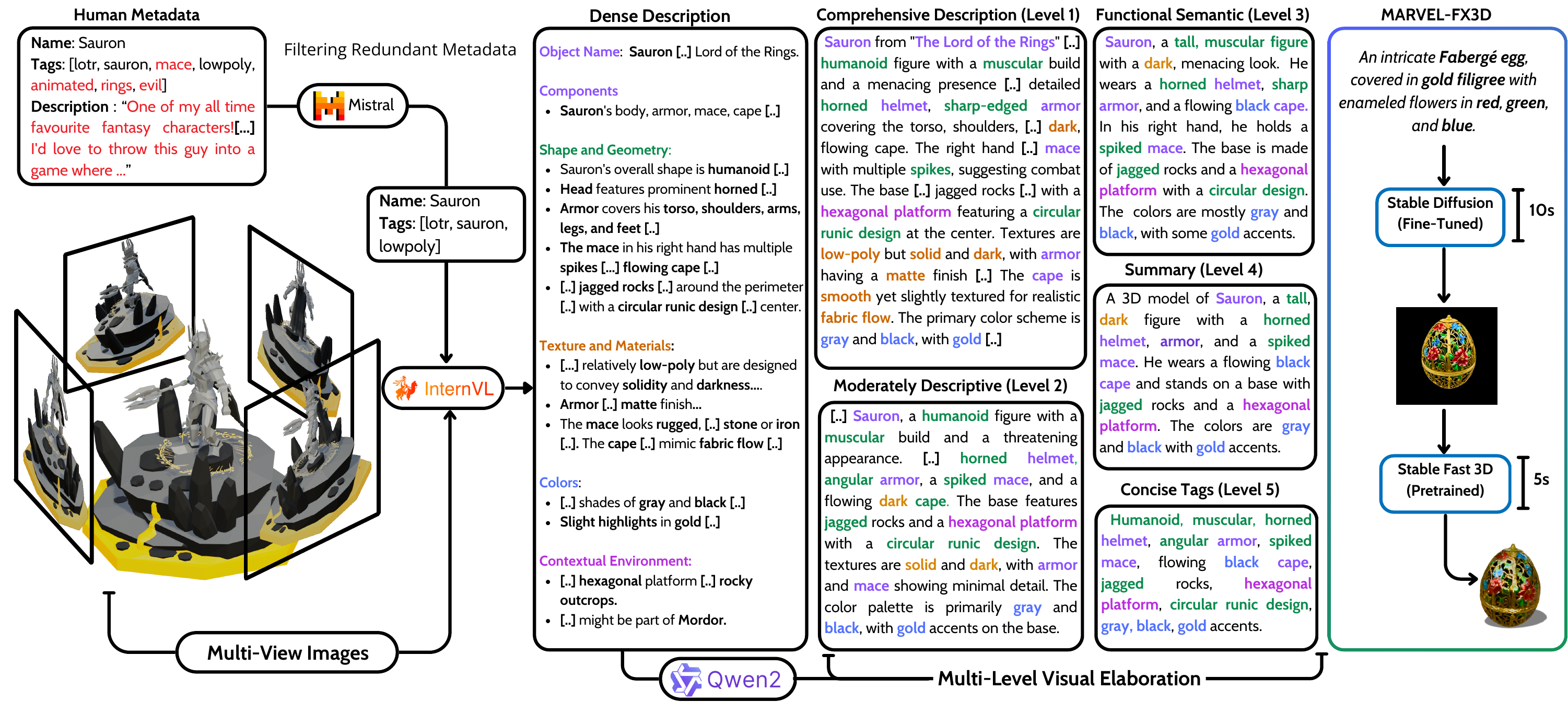}
       \caption{\textbf{Left}: MARVEL annotation pipeline for 3D assets.  Our pipeline starts with human metadata~\cite{objaverse,objaversexl} and rendered multi-view images to create detailed visual descriptions using InternVL-2~\cite{internvl}. These contain \textcolor{annot_violet}{object names}, \textcolor{annot_green}{shapes}, \textcolor{annot_orange}{textures}, \textcolor{annot_blue}{colors}, and \textcolor{annot_pink}{environments}. Qwen2~\cite{qwen2} then processes these descriptions into five hierarchical levels, progressively compressing different aspects of the 3D assets. \textbf{Right}: Our Text-to-3D pipeline finetunes SD 3.5~\cite{sd3,sd3_huggingface} with this dataset and uses pretrained SF3D~\cite{sf3d} to generate a textured mesh in \infertime.}
   \vspace*{-.8\baselineskip}
    \label{fig:annot_pipeline}
    \vspace*{-.6\baselineskip}
\end{figure*}
\subsection{Multi-Stage Annotation Pipeline} \label{annot_pipeline}

We now present our proposed MARVEL annotation pipeline, shown in Figure~\ref{fig:annot_pipeline} (left). Our goal is to generate detailed and domain-specific captions, for both fine-grained and abstract 3D modeling cases. Through a carefully designed five-stage process, MARVEL produces a hierarchy of information-rich and domain-specific annotations. These annotations range from detailed descriptions of object names, shapes, textures, and contextual relationships to concise summaries. Starting with multi-view rendering, our pipeline processes each asset through sequential stages of human metadata refinement, dense description generation via InternVL2~\cite{internvl1.5}, multi-level elaboration using Qwen $2.5$~\cite{qwen2}, and ethical filtering. Below, we detail each component of our pipeline.

\vspace{0.2cm}

\noindent \textbf{Multi-View Rendering:} We first generate $4$ multi-view images of resolution $512\times 512$ for each 3D model using Blender~\cite{blender}. We rotate the camera around the object with azimuth angle, $\theta=\{\frac{\pi i}{2} \}_{i=1}^{i=4}$
and fixed elevation angle, $\phi=30$. Models are scaled to a unit bounding box, centered at the origin, with the camera distance set as $1.5$ relative to this. The four images correspond to the front, back, left, and right sides of the 3D model. Unlike existing 3D captioning pipelines~\cite{cap3d,3dtopia}, we focus solely on these standard viewpoints. This method aligns with recent studies~\cite{ruan2024omniviewtuningboostingviewpointinvariance,woo2024ritualrandomimagetransformations}, which demonstrate that VLMs perform better on images from these viewpoints.
\vspace{0.2cm}
\noindent \textbf{Human Metadata Filtering:} High-quality 3D annotation requires capturing both visual characteristics (\eg. shape, color, texture) and semantic properties (\eg. domain-specific nomenclature and object identification). This dual focus ensures that descriptions are not only visually precise but also contextually relevant within specific domains. A significant challenge in this process is the tendency of pretrained VLMs~\cite{blip2} to hallucinate when dealing with complex datasets, such as Objaverse~\cite{objaverse,objaversexl}, due to the inherent 2D-3D domain gap~\cite{ICML2024_LeveragingVLMs}. To address this, we use the user-generated metadata from source datasets, which provides valuable domain-specific names and descriptions that can guide VLMs~\cite{internvl,internvl1.5} toward generating more precise and informative annotations. However, this metadata often contains noise, including personalized or sensitive information~\cite{objaverse, objaversexl}, which can compromise annotation quality. To mitigate this, we use Mistral-Nemo-Instruct-2407~\cite{nvidianemo} to filter the metadata, removing random, redundant, and sensitive content to ensure that only information relevant to 3D attributes is passed to the annotation pipeline. It is worth noting that our pipeline functions independently of human metadata, with it serving purely as an optional enhancement to add domain-specific terminology in the captions.

\vspace{0.2cm}
\noindent \textbf{Dense Description Generation:} In this stage, InternVL2~\cite{internvl,internvl1.5} takes as input the $4$ rendered multi-view images along with our metadata-augmented prompt to generate a dense description of the 3D models. This step avoids separate processing and view aggregation~\cite{cap3d,3dtopia} required by pipelines using models like LLaVA~\cite{llava,llava_next} or BLIP-2~\cite{blip,blip2}. The generated description contains several key requirements for fine-grained 3D model reconstruction: (1) \textit{structural decomposition} with object identification and relative positions, (2) \textit{geometric properties}, analyzing shape characteristics, symmetry axes, and proportional relationships, (3) \textit{surface characteristics}, addressing texture and material properties and tactile qualities such as roughness and reflectivity; (4) \textit{chromatic analysis}, mapping colors across primary objects and sub-components, including patterns and transitions (5) \textit{environmental context}, capturing spatial relationships and its interaction with other elements.

\noindent To efficiently scale this process for large-scale annotation, we select InternVL2-$40$B~\cite{internvl1.5,internvl} for its balance of speed, accuracy, and prompt adherence. Recent studies show that InternVL2-40B~\cite{internvl1.5,internvl} performs comparably~\cite{internvl,internvl1.5,open_vlm_leaderboard} to GPT-4o~\cite{achiam2023gpt} with significantly lower annotation cost\footnote{InternVL2-$40$B~\cite{internvl1.5,internvl} ranked third on the Huggingface Open-VLM Leaderboard~\cite{open_vlm_leaderboard} during our project.}.

\vspace{0.2cm}
\noindent \textbf{Multi-Level Visual Elaboration:} This stage focuses on generating multi-level visual elaborations using Qwen2.5-$72$B~\cite{qwen2} by compressing different aspects of 3D reconstructions at varying levels of detail. This hierarchical approach allows for flexible and adaptive 3D modeling outputs optimized for different use cases, such as scenarios where only key details—like \textit{object name} and \textit{colors}—are specified, but \textit{texture} is excluded or where simplified \textit{semantic} \textit{tags} is necessary for rapid prototyping. While a direct prompting method will be to specify which aspects to compress, we found that this strategy often constrains the model's ability to create rich and meaningful captions, aligning with findings from recent studies~\cite{letmespeakfreely,white2023prompt}. To overcome these challenges, we develop a hierarchical prompting strategy that specifies the essential content for each level of elaboration, balancing detail and brevity. Below, we describe each level:

\begin{enumerate}
\item \textbf{Comprehensive Description (Level 1):} A detailed description covering all aspects of the 3D model, including precise geometric specifications, materials, spatial relationships, and structural details, in 150-200 words. 


\item \textbf{Moderately Descriptive (Level 2):} Description of the model's primary structures, components, and key geometric features. This level focuses on the overall shape and main features of the model in 100-150 words.


\item \textbf{Functional-Semantic Description (Level 3):} Basic description about the model's functional aspects, general form, and primary characteristics in 50-100 words.

\item \textbf{Summary (Level 4):} A brief description of the object, highlighting its basic form, purpose, and most notable features, similar to existing datasets in 30 words.

\item \textbf{Concise tags (Level 5):} A list of distinct concepts of the 3D model for rapid 3D modeling in 10-20 words.
\end{enumerate}


\noindent An example of multi-level visual elaboration is illustrated in Figure~\ref{fig:annot_pipeline} (left), where Qwen $2.5$~\cite{qwen2} compresses texture information progressively from Level 1 to Level 4. By Level 5, the output shifts to a concise format with colored words representing key semantic tags and core attributes.  To assess the effectiveness of this hierarchical compression, we conduct an ablation study in Section~\ref{subsec:ablation_study}B, measuring the retention of semantic information across all levels.

\vspace{0.2cm}
\noindent \textbf{Ethical Filtering:}
Given the diverse metadata sources in our annotation pipeline, there is a risk of ethically problematic content being included in the multi-level descriptions. To mitigate this, we use the Qwen $2.5$-$14$B~\cite{qwen2} model with a targeted prompt for ethical filtering. This prompt removes meaningless or offensive words, personal names (unless they are famous or contextually relevant), and overly specific identifiers. Importantly, it retains well-known terms, scientific and cultural references, preserving valuable context. This filtering step maintains annotation quality, prevents the leakage of sensitive or inappropriate information, and upholds the integrity of the dataset. For details, please refer to the supplementary material.

\begin{figure*}[t]
    \centering
    \includegraphics[width=1\linewidth]{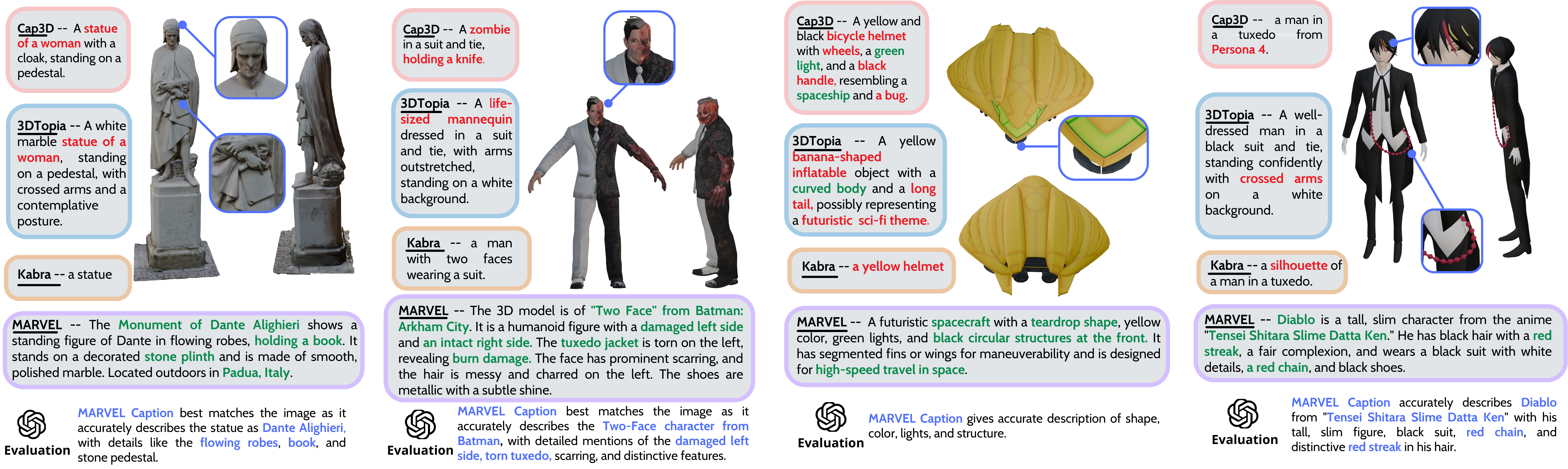}
    \caption{Qualitative Annotation Comparison: From top to bottom Cap3D~\cite{cap3d}, 3DTopia~\cite{3dtopia}, Kabra~\cite{ICML2024_LeveragingVLMs}, MARVEL (Level-4) annotations and GPT-4~\cite{achiam2023gpt} evaluation. MARVEL consistently provides the most comprehensive and precise annotations, capturing intricate details such as object names, color, structure, and specific attributes. \textcolor{red}{Red} is for wrong captions. \textcolor{darkgreen}{Green} shows important information.}
    \vspace*{-.5\baselineskip}
    \label{fig:annot_compare_baseline}
    \vspace*{-.5\baselineskip}
\end{figure*}

\subsection{Caption Generation}

\noindent \textbf{Datasets:} We aggregate $\sim8.9$M 3D assets from seven diverse sources~\cite{objaverse, objaversexl, omniobject3d, pix3d, shapenet, abo, toys4k, gso}.
For human metadata injection, we use the \textit{name}, \textit{tags} and \textit{description} from Objaverse 1.0~\cite{objaverse} (Sketchfab) and \textit{metadata} from Objaverse-XL~\cite{objaverse} (thingiverse and github). Samples from Objaverse-XL~\cite{objaversexl} containing the file extension \textit{.ply} were excluded from the dataset. For the rest of the six datasets~\cite{shapenet,abo,gso,omniobject3d,toys4k,pix3d}, we use the class categories as metadata. Samples without any renderable multi-view images or contains zero-length final annotations are removed from the dataset. The final details of the dataset are provided in Table~\ref{tab:datasets_overview}, with additional preparation information available in the supplementary materials.

\vspace{0.1cm}
\noindent \textbf{Implementation Details:} 
Our MARVEL annotation pipeline is optimized for large-scale processing, achieving a throughput of $\sim$$24,000$ samples per day. 
For human metadata filtering, we run the Mistral-Nemo-Instruct-2407~\cite{nvidianemo} on a single NVIDIA RTX $4090$ GPU. Both InternVL2-$40$B~\cite{internvl,internvl1.5} for dense description generation and the Qwen$2.5$-$72$B~\cite{qwen2} with 8-bit quantization for multi-level visual elaboration, runs on a single NVIDIA H$100$ GPU. For the final ethical filtering stage, we run Qwen $2.5$-$14$B~\cite{qwen2} on a single NVIDIA RTX A$6000$ GPU. For complete details including hyperparameter details, please refer to the supplementary material.

\subsection{MARVEL-FX3D Architecture}\label{marvel_fx3d_architecture}

In this section, we present MARVEL-FX3D, a two-stage pipeline that demonstrates the practicality of the MARVEL-$40$M+ dataset for TT3D synthesis. By leveraging our dataset's comprehensive text descriptions and diverse 3D assets~\cite{objaverse}, MARVEL-FX3D generates high-quality textured 3D meshes from text descriptions that can specify multiple objects, scenes, geometric properties, colors, and textures. The pipeline consists of (1) TTI generation using fine-tuned Stable Diffusion~\cite{sd3}, followed by (2) single-view 3D reconstruction with a pretrained view reconstruction model~\cite{sf3d}. This entire process generates high quality 3D assets in \infertime, as illustrated in Figure~\ref{fig:annot_pipeline} (right).

\vspace{0.1cm}
\noindent \textbf{Fine-Tuning TTI Model:} The objective of this stage is to generate high-quality, diverse images from text prompts that can be effectively converted into 3D textured meshes using pretrained image-to-3D methods~\cite{instantmesh,sf3d}. A primary challenge in multi-stage TT3D pipelines~\cite{pointe,dual3d,assetgen,xdreamer,instant3d} is the inherent 2D-3D domain gap, where reconstructing accurate and geometrically consistent 3D models from 2D images is hindered by the ambiguity between background and foreground information~\cite{dual3d,xdreamer}. To address this, some methods have fine-tuned TTI \cite{StableDiffRombach22,deepfloydif} models on TT3D datasets~\cite{cap3d,zhang2024clay,3dtopia,ICML2024_LeveragingVLMs}. Following this approach, we fine-tune Stable Diffusion $3.5$~\cite{sd3,sd3_huggingface} using the LORA \cite{hu2021lora} strategy to bridge this domain gap and generate images similar to the training distribution of the image-to-3D methods~\cite{sf3d}.


\vspace{0.1cm}


\noindent \textbf{Image-to-3D Generation:}  In the second stage, the background is removed from the generated image using DIS~\cite{backgroundRemoval}. The refined image is then passed to pretrained SF3D~\cite{sf3d} to generate a high-quality textured mesh in 5s.


\vspace{0.2cm}




\vspace*{-.6\baselineskip}
\section{Experiment} \label{experiment}
The experiment section is divided into two parts. In Sec.\ref{subsec:annot_qual}, we evaluate the quality of our annotations in comparison to the baseline datasets \cite{cap3d,3dtopia,ICML2024_LeveragingVLMs}. While Sec.~\ref{subsec:baseline} presents the performance evaluation of MARVEL-FX3D against current state-of-the-art methods \cite{dreamfusion,shap_e,luciddreamer,hifa}. Both experiments are conducted on Objaverse \cite{objaverse} dataset.

\subsection{Annotation Evaluation} \label{subsec:annot_qual}

\noindent \textbf{Experimental Setup and Metrics:} We assess annotation quality through (1) Linguistic Assessment, (2) Image-Text Alignment, and (3) Caption Accuracy.

\noindent The linguistic assessment evaluates annotation richness and diversity using the Measure of Textual Lexical Diversity (MTLD) \cite{mtld} and N-gram analysis \cite{brown1992estimate}. The MTLD metric calculates the average segment length at which the type-token ratio (TTR) drops below a threshold (typically $0.72$), with higher MTLD scores indicating more diverse annotations. We randomly select $50$K annotations for analysis. 

\vspace{0.1cm}
\noindent Image-text alignment is measured using both GPT-4 \cite{achiam2023gpt} and human evaluators who review four multi-view images of each 3D model and select the best-matching caption. $5,000$ samples are evaluated using GPT-4 and $1,000$ samples by five human reviewers with each assigned $200$ samples. Level 4 annotations from MARVEL-40M+ are used for fair comparison due to their similar average length to baseline datasets~\cite{cap3d,3dtopia,ICML2024_LeveragingVLMs} as shown in Table~\ref{tab:linguistic analysis}.

\noindent Caption accuracy is separately assessed, where GPT-4 and human reviewers evaluate whether all the 3D attributes mentioned in the captions accurately correspond to the 3D models using four multi-view images. For MARVEL-40M+, Level 1 annotations are used, which are detailed and form the foundation for subsequent levels. GPT-4 evaluates $1,000$ samples, while human reviewers assess $250$ samples due to the evaluation's time demands. More details are provided in the supplementary section.



\begin{table}[h]
\centering
\resizebox{\columnwidth}{!}{%

\begin{tabular}{lcccc|cc}
\hline
\vspace*{-.6\baselineskip}
Dataset  & \begin{tabular}[c]{@{}l@{}}Average \\ Length\end{tabular} & \begin{tabular}[c]{@{}l@{}}MTLD \cite{mtld} \\(@50K)\end{tabular} & \begin{tabular}[c]{@{}l@{}}Unigram \\(@50K)\end{tabular} & \begin{tabular}[c]{@{}l@{}}Bi-Gram \\(@50K)\end{tabular} & \begin{tabular}[c]{@{}l@{}}GPT4 \\(@5K)\end{tabular} & \begin{tabular}[c]{@{}l@{}}Human \\(@1K)\end{tabular} \\
\hline
Cap3D \cite{cap3d}    & 16                                                      & 39.71                                                   & 15,189 & 123,071 & 14.55                                  & 9.50                             \\
3D-Topia \cite{3dtopia} & 29                                                    & 41.43                                                   & 10,329 & 95,856  & 10.80                                  & 14.00                            \\
Kabra \cite{ICML2024_LeveragingVLMs} & 5                                      & 25.85                                                   & 3,862  & 19,753  &   2.24                                     &   3.10                               \\
MARVEL (Level 4)   & \textbf{44}                                            & \textbf{47.43}                                          & \textbf{27,659} & \textbf{239,052} & \textbf{72.41}                        & \textbf{73.40}                   \\
\hline
\end{tabular}%
}
\caption{Quantitative comparison of annotation quality across datasets. MARVEL surpasses existing datasets \cite{ICML2024_LeveragingVLMs,3dtopia,cap3d} in all metrics, showcasing superior linguistic diversity, vocabulary coverage, and significantly higher ratings from GPT-4 and humans.}
\vspace*{-.6\baselineskip}
\label{tab:linguistic analysis}
\vspace*{-.5\baselineskip}
\end{table}

\vspace{0.1cm}
\noindent \textbf{Linguistic Assessment:} Table \ref{tab:linguistic analysis} (left) shows the MTLD \cite{mtld} score and N-Gram analysis \cite{brown1992estimate}. MARVEL demonstrates a notable improvement, achieving an MTLD score approximately $83\%$ higher than Kabra \cite{ICML2024_LeveragingVLMs} $19$\% higher than Cap3D \cite{3dtopia} and $14$\% higher than 3D-Topia\cite{cap3d} signifying richer caption diversity.  In addition, MARVEL shows a significantly higher unigram vocabulary size, surpassing Kabra \cite{ICML2024_LeveragingVLMs}, Cap3D \cite{cap3d} and 3D-Topia \cite{3dtopia}  by factors of approximately $7.1$, $1.8$ and $2.6$ respectively. The trend extends to bigram analysis and average word length as well, showing MARVEL's superior linguistic diversity and information density. Figure~\ref{fig:annot_compare_baseline} also illustrates that MARVEL’s annotations contain more unique words, particularly focusing on object names, colors, textures, and attributes.

\vspace{0.1cm}
\noindent \textbf{Image-Text Alignment:} As shown in Table~\ref{tab:linguistic analysis}, MARVEL achieves notably higher ratings in image-text alignment, with win rates of $72.41\%$ from GPT-4 \cite{gpt4_eval_human} and $73.40\%$ from human evaluators, outperforming prior methods. This reflects MARVEL's superior alignment of captions with 3D models. Figure \ref{fig:annot_compare_baseline} highlights this through examples, showing that MARVEL's detailed descriptions capture nuances, such as the \textquote{\textit{flowing robes and book-holding posture of a historical statue}}, even at Level 4. In contrast, baselines \cite{ICML2024_LeveragingVLMs,3dtopia,cap3d} produce simpler, more generic descriptions. MARVEL's integration of filtered human-annotated metadata further enhances the identification of complex, domain-specific entities like \textquote{\textit{Monument of Dante Alighieri}}, \textquote{\textit{Two Face}}, and \textquote{\textit{Diablo}}.


\vspace{0.1cm}
\noindent \textbf{Caption Accuracy:} Table~\ref{tab:caption_consistency} shows the caption accuracy results, where MARVEL (Level 1) achieves the highest scores—$84.70\%$ in GPT-4 evaluation and $82.80\%$ in human evaluation—demonstrating superior consistency compared to other methods. Baseline datasets with shorter captions, like Kabra~\cite{ICML2024_LeveragingVLMs} and others~\cite{3dtopia,cap3d}, tend to capture objects semantically (e.g., \textquote{\textit{a statute}},\textquote{\textit{a man in a tuxedo}}) but lack detailed descriptions. Although longer captions increase the risk of errors, MARVEL (Level 1) maintains high accuracy with an average length of 170 words ($34\times$ that of Kabra \cite{ICML2024_LeveragingVLMs}, $10\times$ that of Cap3D \cite{cap3d}, and $5\times$ that of 3D-Topia \cite{3dtopia}), effectively balancing detail and correctness. As shown in Figure~\ref{fig:annot_compare_baseline}, MARVEL captures both domain-specific names and intricate features, exemplified by captions like \textquote{\textit{The Monument of Dante Alighieri... flowing robes, holding a book ... smooth, polished marble...}}

\begin{table}[ht]
\vspace*{-.6\baselineskip}
\resizebox{\columnwidth}{!}{
\begin{tabular}{lccc}
\hline
\multicolumn{1}{l}{\multirow{2}{*}{Method}} & \multirow{2}{*}{\begin{tabular}[c]{@{}c@{}}\\ Average Length\end{tabular}} & \multicolumn{2}{c}{Correct} \\
\multicolumn{1}{l}{} &  & \begin{tabular}[c]{@{}c@{}}GPT4 Evaluation \\ (@1k)\end{tabular} & \begin{tabular}[c]{@{}c@{}}Human Evaluation\\ (@250)\end{tabular} \\ \hline
Cap3D \cite{cap3d} & 16 & 76.00 & 72.80 \\
3D-Topia \cite{3dtopia} & 29 & 54.60 & 44.80 \\
Kabra \cite{ICML2024_LeveragingVLMs} & 5 & 83.40 & 78.20 \\
MARVEL (Level 1) & 170 & \textbf{84.70} & \textbf{82.80} \\ \hline
\end{tabular}
}
\vspace*{-.4\baselineskip}
\caption{Comparison of caption accuracy using GPT-4V \cite{achiam2023gpt} and humans, highlighting MARVEL's (Level 1) superior consistency desite significantly higher caption length.}
\vspace*{-.3\baselineskip}
\label{tab:caption_consistency}
\end{table}




\begin{figure}[t]
    \centering
    \includegraphics[width=0.96\linewidth]{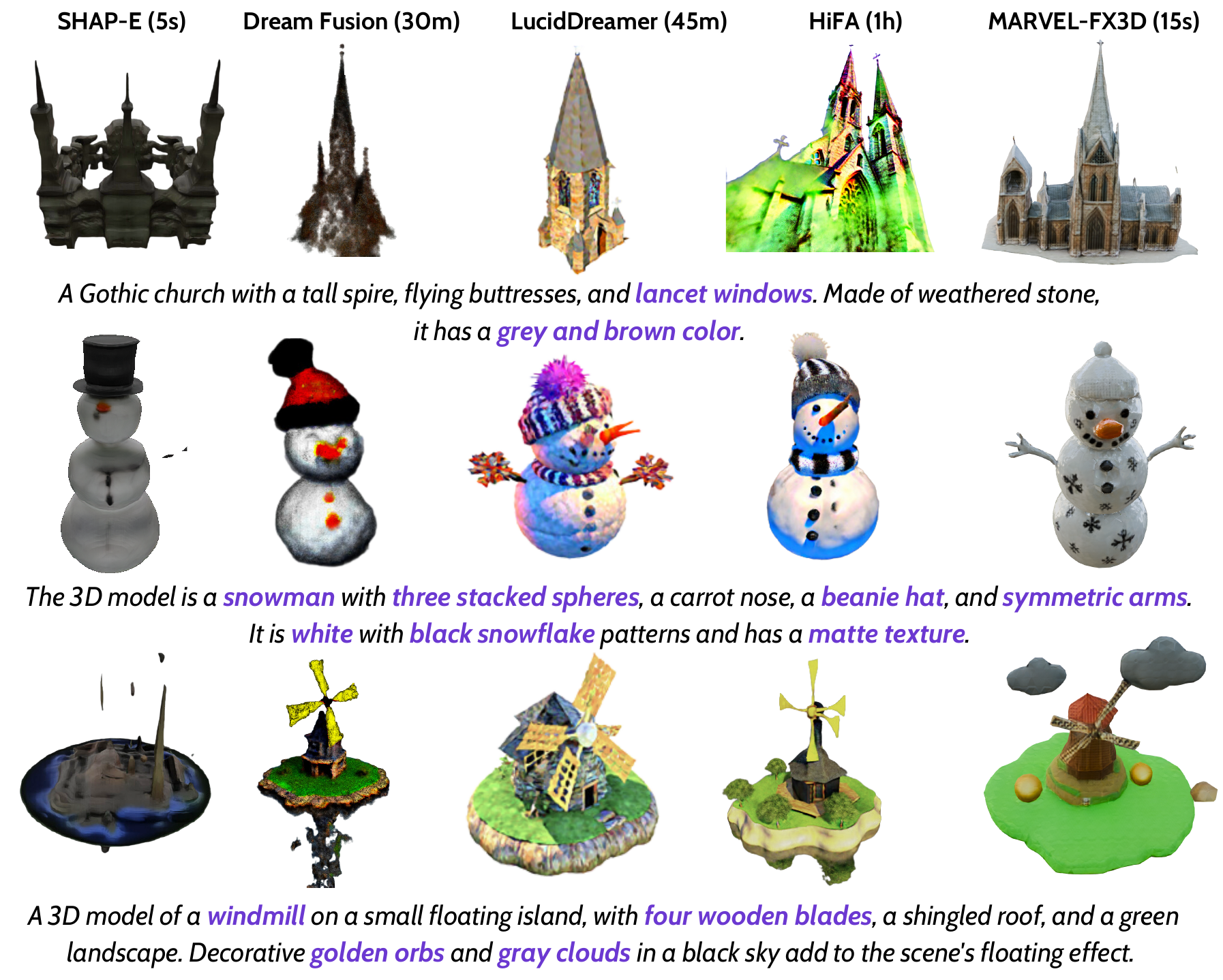}
    \caption{Visual results of high fidelity TT3D generation. Left to right, the reconstructed 3D assets from Shap-E \cite{shap_e}, DreamFusion \cite{dreamfusion}, Lucid-Dreamer \cite{luciddreamer}, HIFA \cite{hifa} and MARVEL-FX3D.}
    \vspace*{-.5\baselineskip}
    \label{fig:qual_baselines}
    \vspace*{-.7\baselineskip}
\end{figure}

\vspace*{-.5\baselineskip}
\subsection{Text-to-3D Generation}\label{subsec:baseline}

\noindent \textbf{Implementation Details:} We fine-tune SD $3.5$ \cite{sd3,sd3_huggingface} using the Objaverse \cite{objaverse} dataset, which includes $798,759$ 3D assets, split into training, validation, and test sets in a $90:5:5$ ratio. Fine-tuning is conducted in half-precision for $5$ epochs with a batch size of $8$, using a single NVIDIA H$100$ GPU, with LoRA \cite{hu2021lora} rank and alpha set to $4$.


\vspace{0.1cm}

\noindent \textbf{Baselines:} To assess MARVEL-FX3D's performance in high-fidelity TT3D generation, we compare it with Shap-E~\cite{shap_e}, Dreamfusion~\cite{dreamfusion}, Luciddreamer~\cite{luciddreamer}, and HIFA~\cite{hifa}. Due to their slower optimization~\cite{dreamfusion,luciddreamer,hifa}, we evaluate 50 random level 4 captions from the Objaverse \cite{objaverse} test set. Instant3D~\cite{instant3d} and Assetgen~\cite{assetgen} are excluded due to unavailable code. See supplementary for details.


\vspace{0.1cm}

\noindent \textbf{User Study:} We conducted a human evaluation to assess the geometric consistency, visual quality, prompt fidelity, and overall preference of reconstructed 3D assets. Geometric consistency measures realism and physical plausibility, identifying issues like the \textit{janus problem}. Prompt fidelity evaluates alignment with input text, while visual quality considers aesthetic elements such as colors and textures. Five users were presented with the text prompt and videos of the rendered 3D assets generated by the baseline methods and MARVEL-FX3D. The users scored each asset separately from 1 to 10 based on these criteria, and the final scores were averaged across all users.

\begin{table}[ht]
\centering
\resizebox{\columnwidth}{!}{%
\begin{tabular}{lccccc}
\hline
\multirow{2}{*}{} & \multirow{2}{*}{Time$\downarrow$} & Geometric  & Visual & Prompt  & Overall    \\
                  &                                   & Consistency & Quality & Fidelity &  \\ \hline
Shap-E \cite{shap_e}             & \textbf{5s}                                &             3.31$\pm$ 0.71&        2.25 $\pm$ 0.43&           2.65 $\pm$ 0.51&            2.41 $\pm$ 0.50\\
DreamFusion \cite{dreamfusion}   & 30m                               &             4.88 $\pm$ 0.47&        3.74 $\pm$ 0.80&           4.22 $\pm$ 0.79&            4.09 $\pm$ 0.81\\
HiFA \cite{hifa}                 & $>$1h                             &             6.59 $\pm$ 0.57&        6.42 $\pm$ 0.26 &           6.88 $\pm$ 0.46&            6.44 $\pm$ 0.35\\
Lucid-Dreamer \cite{luciddreamer}& 45m                               &             \textbf{7.25} $\pm$ 0.60&        6.47 $\pm$ 1.24&            6.62 $\pm$ 1.37&            6.59 $\pm$ 0.86\\
\hline
MARVEL-FX3D 
                  & 15s                                &             7.20 $\pm$ 0.91&        \textbf{6.58} $\pm$ 0.86 &           \textbf{7.71} $\pm$ 0.68&            \textbf{6.94} $\pm$ 0.71\\ \hline
\end{tabular}
}
\vspace*{-.6\baselineskip}
\caption{Quantitative evaluation focusing on time and human evaluation criteria: geometric consistency, visual quality, prompt fidelity, and overall preference.}
\label{tab:baseline_comparison}
\vspace*{-.4\baselineskip}
\end{table}

\noindent \textbf{Results:} Table~\ref{tab:baseline_comparison} presents the quantitative comparison of MARVEL-FX3D against the baselines \cite{dreamfusion,luciddreamer,hifa,shap_e}. MARVEL-FX3D shows notable improvements, achieving the highest prompt fidelity ($7.71$) and overall preference ($6.94$), indicating strong alignment with input descriptions and balanced performance across criteria. It also tops visual quality with a score of $6.58$, slightly ahead of Lucid-Dreamer \cite{luciddreamer} ($6.47$), which marginally exceeds MARVEL-FX3D in geometric consistency ($7.25$ vs. $7.20$) due to occasional flat outputs from SF3D \cite{sf3d}. Despite this, MARVEL-FX3D's processing time is significantly faster, completing in just \infertime compared to Lucid-Dreamer's $45$ minutes, HiFA's over $1$ hour, and DreamFusion's $30$ minutes. Shap-E \cite{shap_e}, while the quickest at $5$ seconds, shows considerably lower performance across all metrics. Figure~\ref{fig:qual_baselines} includes some qualitative examples.

\vspace{0.1cm}
\noindent \textbf{Cap3D vs MARVEL-40M+:} To demonstrate the effectiveness of our dataset on TT3D generation, we conduct human evaluation on the performance of MARVEL-FX3D when SD $3.5$ is either pretrained or finetuned on MARVEL-40M+ and Cap3D~\cite{cap3d} captions. The evaluation follows the same strategy and uses the same 50 test samples as before. Results on Table~\ref{tab:cap3dvsmarvel} confirm that MARVEL-FX3D trained on our captions consistently outperforms pretrained and Cap3D versions across all metrics, with a notable increase in prompt fidelity.\\

\begin{table}[ht]
\vspace{-0.5cm}
\centering
\resizebox{1\linewidth}{!}{
\begin{tabular}{l|c|cccc}
\hline
\multirow{2}{*}{Dataset} & \multirow{2}{*}{Text-to-Image~\cite{sd3_huggingface}} & Geometric  & Visual & Prompt  & Overall    \\
                  &                                    & Consistency & Quality & Fidelity &  \\ \hline
\rule{1cm}{1pt}  & Pretrained & 2.51 ± 1.84 & 2.54 ± 1.78 & 2.58 ± 1.86 & 2.41 ± 1.69 \\
Cap3D  & Finetuned & 6.51 ± 1.48 & 6.53 ± 1.52 & 6.54 ± 1.68 & 6.43 ± 1.47 \\
MARVEL & Finetuned & \textbf{7.20 ± 0.91} & \textbf{6.58 ± 0.86} & \textbf{7.71 ± 0.68} & \textbf{6.94 ± 0.71} \\ \hline
\end{tabular}
}

\caption{Quantitative evaluation of MARVEL-FX3D without and with fine-tuning on Cap3D and MARVEL captions, showing improved performance with MARVEL.}
\label{tab:cap3dvsmarvel}
\vspace*{-.4\baselineskip}
\end{table}

\vspace*{-.7\baselineskip}
\subsection{Ablation Study} \label{subsec:ablation_study}

\begin{figure}[t]
    \centering
    \includegraphics[width=1\linewidth]{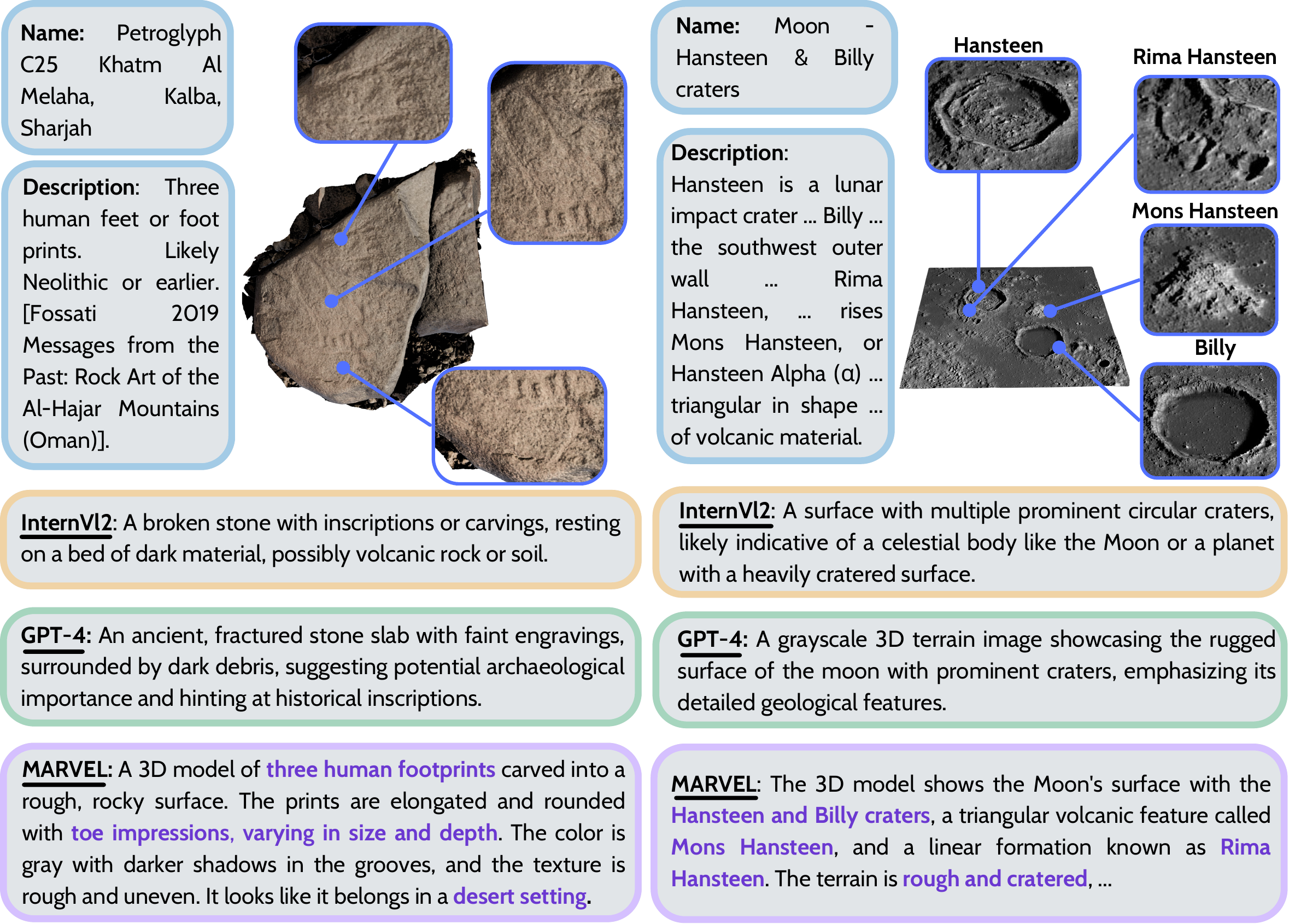}
    \caption{MARVEL uses human-generated metadata from source datasets to create detailed, accurate captions (e.g., names of the \textit{lunar craters}, detection of \textit{human footprints}) and reduce hallucinations. Without metadata, VLMs like GPT-4 \cite{achiam2023gpt} and InternVL2 \cite{internvl1.5} generate vague or speculative descriptions.}
    \vspace*{-.5\baselineskip}
    \label{fig:metadata}
    \vspace*{-.5\baselineskip}
\end{figure}

\vspace{0.1cm}
\noindent \textbf{A. Effect of Human Metadata on Annotation Quality:} Human-generated metadata is vital in the MARVEL annotation pipeline, enriching text captions with domain-specific details. While quantitative analysis requires specialized expertise, we provide its qualitative evidence. As shown in Figure~\ref{fig:metadata}, MARVEL accurately identifies specific details, such as \textquote{\textit{three human footprints on a rocky surface}}, which both InternVL-2 \cite{internvl,internvl1.5} and GPT-4 \cite{achiam2023gpt} miss, producing only generic descriptions. Similarly, MARVEL captures identifiers like specific lunar craters, absent in InternVL-2 and GPT-4 outputs. This highlights how integrating human metadata enhances the context in annotations. Additional examples from other domains (e.g., biology, historical sites) are in the supplementary material.

\vspace{0.1cm}
\noindent \textbf{B. Inter-Level Semantic Retention Evaluation:} This ablation study measures how well semantic information is retained across MARVEL-$40$M+ annotation levels as they progress from detailed descriptions to concise tokens. To evaluate this, we report the semantic similarity (cosine similarity of embeddings) between levels using sentence-BERT \cite{reimers2019sentence} and the compression ratio (word count ratio) \cite{compression_ratio}. Results in Table~\ref{tab:SCS} show strong semantic retention from Levels 1-4, demonstrating effective compression while preserving meaning. However, the shift to Level 5 results in lower similarity, reflecting the transition to a list of concepts at the expense of cohesive descriptions.

\begin{table}[H]
\vspace*{-.5\baselineskip}
\centering
\resizebox{0.85\columnwidth}{!}{%
\begin{tabular}{llll}
\hline
\begin{tabular}[c]{@{}l@{}}Source Level\end{tabular} & \begin{tabular}[c]{@{}l@{}}Target Level\end{tabular} & \begin{tabular}[c]{@{}l@{}}Semantic\\ Similarity \end{tabular} & \begin{tabular}[c]{@{}l@{}}Compression\\ Ratio\end{tabular} \\ \hline
Level 1                               & Level 2          
                                      &                       0.91                                &                            0.30                                      \\
Level 2                                                & Level 3                                                &                             0.92                                      &               0.27                                                   \\
Level 3                                                & Level 4                                                &                                 0.88                                  &                 0.47                                                 \\
Level 4                                                & Level 5                                                &                                 0.72                                  &                  0.22                                                \\ \hline
\end{tabular}}
\vspace*{-.5\baselineskip}
\caption{MARVEL-40M+ illustrates strong semantic retention through Levels 1-4 and reduced detail at Level 5.}
\label{tab:SCS}
\vspace*{-.8\baselineskip}
\end{table}

\vspace*{-.8\baselineskip}
\section{Limitation}

%
Our analysis reveals some limitations in MARVEL annotation pipeline and MARVEL-FX3D. First, the underlying VLMs and LLMs struggle with numerical precision~\cite{parcalabescu2021seeingpastwordstesting, campbell2024understandinglimitsvisionlanguage} and directional understanding~\cite{hoehing2023whatsleftcantright} in complex scenes with multiple objects and occlusion. Second, InternVL-2 misidentifies very thin objects, often treating side-views as separate entities. Finally, without metadata, captions become generic for complex 3D structures, especially in fragmented geometries like architectural interiors.  In supplementary, some visual examples are provided. Additionally, MARVEL-FX3D sometimes generates flat 3D objects due to depth ambiguity in the input image. Despite these challenges, our pipeline remains model-agnostic and adaptable to future enhancements.


\vspace*{-.4\baselineskip}
\section{Conclusion}

We present MARVEL-$40$M+, the largest 3D captioning dataset to date, comprising over $40$M+ high-quality text annotations for $8.9$ million 3D assets across seven major 3D datasets. Our primary contributions include a scalable, multi-stage annotation pipeline that combines open-source pretrained multi-view VLMs and LLMs with filtered human metadata to reduce hallucinations and introduce domain-specific information. Our pipeline produces five levels of annotations for diverse 3D modeling needs, from detailed descriptions to concise tags. Additionally, we introduce MARVEL-FX3D, a two-stage architecture that leverages fine-tuned Stable Diffusion on our dataset and pretrained Stable Fast 3D to generate high-quality, textured 3D meshes in just $15$s. Through extensive experimentation, we demonstrated both MARVEL-$40$M+'s superior annotation quality and linguistic depth, and MARVEL-FX3D's improved performance in high fidelity TT3D generation.  We believe MARVEL-$40$M+ will be a foundational resource for advancing TT3D content creation, inspiring further research to address current limitations and expand its applications.



\section{Acknowledgement}

This work was co-funded by the European Union under Horizon Europe, grant number $101135724$, project LUMINOUS. However, the views and opinions expressed are those of the author(s) only and do not necessarily reflect those of the European Union. Neither the European Union nor the granting authority can be held responsible. Human evaluation for this project was primarily funded by BITS Pilani Hyderabad's NFSG Grant (Reference N4/24/1033).
{
    \small
    \bibliographystyle{ieeenat_fullname}
    \bibliography{main}
}

\twocolumn[{%
\renewcommand\twocolumn[1][]{#1}%
\maketitlesupplementary
\begin{center}
    \centering
    \captionsetup{type=figure}
    \includegraphics[width=0.9\linewidth]{supp/assets/supp_teaser.pdf}
    \captionof{figure}{ An example use case of MARVEL-FX3D, demonstrating how multiple prompts can be combined to create a detailed and complex 3D scene, with each prompt contributing specific elements such as characters, structures, and environmental details (Zoom in for details).}
       \label{fig:supp_teaser}
       \vspace{0.5cm}
\end{center}%
}]


\noindent This supplementary material provides additional details and results to support the main paper. Section~\ref{sec:dataprep} outlines the captioning process, including dataset preparation and implementation specifics. Sections \ref{sec:annotation} and \ref{sec:tt3d} delve deeper into MARVEL annotations and MARVEL-FX3D results, offering more examples, discussions, and insights into their applications and limitations.
\section{Additional Details on Captioning Process} \label{sec:dataprep}

\subsection{Dataset Preparation}

\noindent \textbf{Objaverse:} Objaverse\footnote{\url{https://objaverse.allenai.org/objaverse-1.0}}~\cite{objaverse} contains $798,759$ 3D assets, with metadata (e.g., \textit{ name, tags, description}) available for $\sim$$93\%$ samples after filtering. From ObjaverseXL~\cite{objaversexl}, we rendered $8,031,637$ assets, of which $\sim$$3.7$M included metadata. After filtering, around $3$M samples are retained as valid metadata.

\noindent \textbf{ShapeNet:}
For the ShapeNet dataset, which contains 52,472 samples, we use the ShapeNet taxonomy as its metadata (e.g., \textit{airplane}, \textit{bowl}, \textit{cap}, \textit{clock}, etc.).

\noindent \textbf{Pix3D:}
For the Pix3D\footnote{\url{http://pix3d.csail.mit.edu/}}\cite{pix3d} dataset, which contains 374 samples, we use the associated category tag as its metadata (e.g., \textit{bed}, \textit{table}, \textit{desk}, \textit{chair}, etc.).

\noindent \textbf{OmniObject3D:}
The Omni-Object-3D\footnote{\url{https://omniobject3d.github.io/}}\cite{omniobject3d} dataset, which contains 5,878 samples, we use the folder names (e.g., \textit{bed}, \textit{table}, \textit{desk}, \textit{chair}, etc.) as our metadata.

\noindent \textbf{Toys4K:}
For the Toys4K\footnote{\url{https://github.com/rehg-lab/lowshot-shapebias/tree/main/toys4k}}\cite{toys4k} dataset, which contains 4,000 samples, we use the folder names (e.g., \textit{car}, \textit{airplane}, \textit{train}, \textit{robot}, etc.) as our metadata.

\noindent \textbf{GSO:} The GSO (Google Scanned Objects)\footnote{\url{https://goo.gle/scanned-objects}}\cite{gso} dataset, which contains 1,030 samples, we use the folder names (e.g., \textit{lamp}, \textit{sofa}, \textit{vase}, \textit{refrigerator}, etc.) as our metadata.

\noindent \textbf{ABO:} The ABO (Amazon Berkeley Objects)\footnote{\url{https://amazon-berkeley-objects.s3.amazonaws.com/index.html}}\cite{abo} dataset, which contains 7,953 samples, provides metadata through listings information. Since these listings are multilingual, we first use the \texttt{nllb-200}\footnote{\url{https://huggingface.co/facebook/nllb-200-distilled-600M}} model to translate the listings to English. The translated English listings are then used as our metadata.

\subsection{Implementation Details} 
For human metadata filtering, we use the Mistral-Nemo-Instruct-2407 model with a temperature of 0.3 and a top-p value of 0.95. For dense description generation, we employ InternVL2-40B, configured with a temperature of 0.70, a top-p value of 0.95, and a repetition penalty of 1.10, with multinomial sampling enabled. For multi-level visual elaboration, we utilize Qwen2.5-72B with 8-bit quantization, a temperature of 0.70, a top-p value of 0.80, and a repetition penalty of 1.05. Finally, the Qwen2.5-14B model, used for the ethical filtering stage, is configured with a temperature of 0 and a top-p value of 0.90.

\noindent For human evaluations in our paper, we developed a Gradio app to compare our captions with those from baseline datasets, including Cap3D, 3DTopia, and Kabra, as well as to evaluate FX3D results against text-to-3D baselines. The evaluations were conducted by a panel of 5 human experts.

\subsection{Compute and GPU Hours} 
MARVEL's annotation pipeline utilizes one NVIDIA H100-80GB GPU, one RTX-4090 GPU, and one RTX-A6000 GPU, achieving a throughput of approximately 24,000 samples per day. Annotating the entire Objaverse dataset (800,000 samples) would thus require about 33 days, incurring an estimated total computational cost of approximately \$2,700–\$3,000, based on publicly available GPU \texttt{pricing}\footnote{\url{https://tinyurl.com/gpu-usage-pricing} (\href{https://cloud.vast.ai/?utm_source=google_ads&utm_medium=circleclick.com&utm_source=google&utm_medium=cpc&utm_campaign=20740487654_&utm_target=&utm_group=&utm_content=&placement=&device=c&adposition=&gad_source=1&gclid=Cj0KCQjwhYS_BhD2ARIsAJTMMQbrq183XBuy-9v05-QeDXInWksFJfOpm564rFPtAQeuVQj6IXF-BcEaAvwoEALw_wcB}{Original})}.


\noindent For comparison, sequential human annotation has a considerably lower throughput (1,400 samples/day) and higher cost (\$87.18 per 1,000 annotations), resulting in approximately 572 days (about 1.57 years) and a total cost of roughly \$69,744 for annotating the complete Objaverse dataset. In contrast, the automated Cap3D pipeline—leveraging BLIP2, CLIP, and GPT-4 models on cloud-hosted NVIDIA A40 GPUs—achieves significantly higher throughput (65,000 samples/day) at a lower cost (\$8.35 per 1k annotations), requiring only about 13 days and totaling approximately \$6,680 for the entire dataset~\cite{cap3d}.

\noindent Our pipeline annotates the Objaverse dataset at approximately half the total cost of Cap3D, although with a lower throughput (33 days vs. Cap3D's 13 days). Both automated methods substantially outperform sequential human annotation in terms of speed and cost. Importantly, our pipeline delivers annotations of significantly higher quality compared to Cap3D, making it particularly advantageous when balancing annotation quality and cost efficiency. All comparisons assume sequential (non-parallelized) processing; parallelization would further reduce annotation time for all methods.

\begin{table}[htbp]
\centering
\resizebox{\columnwidth}{!}{%
\begin{tabular}{lcccc}
\hline
\vspace*{-.6\baselineskip}
Method & \begin{tabular}[c]{@{}c@{}}Throughput\\(samples/day)\end{tabular} & \begin{tabular}[c]{@{}c@{}}Total Days\\(800k samples)\end{tabular} & \begin{tabular}[c]{@{}c@{}}Cost per 1k\\annotations\end{tabular} & \begin{tabular}[c]{@{}c@{}}Total Cost\\(800k samples)\end{tabular} \\ 
\hline
Human & 1,400 & 572 & \$87.18 & \$69,744 \\[2pt]
Cap3D & 65,000 & 13 & \$8.35 & \$6,680 \\[2pt]
MARVEL & 24,000 & 33 & \$3.38–\$3.75 & \$2,700–\$3,000 \\[2pt]
\hline
\end{tabular}%
}
\caption{Comparison of annotation pipelines based on throughput, annotation time, and cost for annotating the Objaverse dataset (800k samples). All estimates assume sequential annotation without parallelization.}
\vspace*{-.6\baselineskip}
\label{tab:annotation-comparison}
\vspace*{-.5\baselineskip}
\end{table}

\section{Additional details on MARVEL annotations} \label{sec:annotation}

\subsection{More Results on Effects of Human Metadata}

\begin{figure*}[h]
    \centering
    \includegraphics[width=1\linewidth]{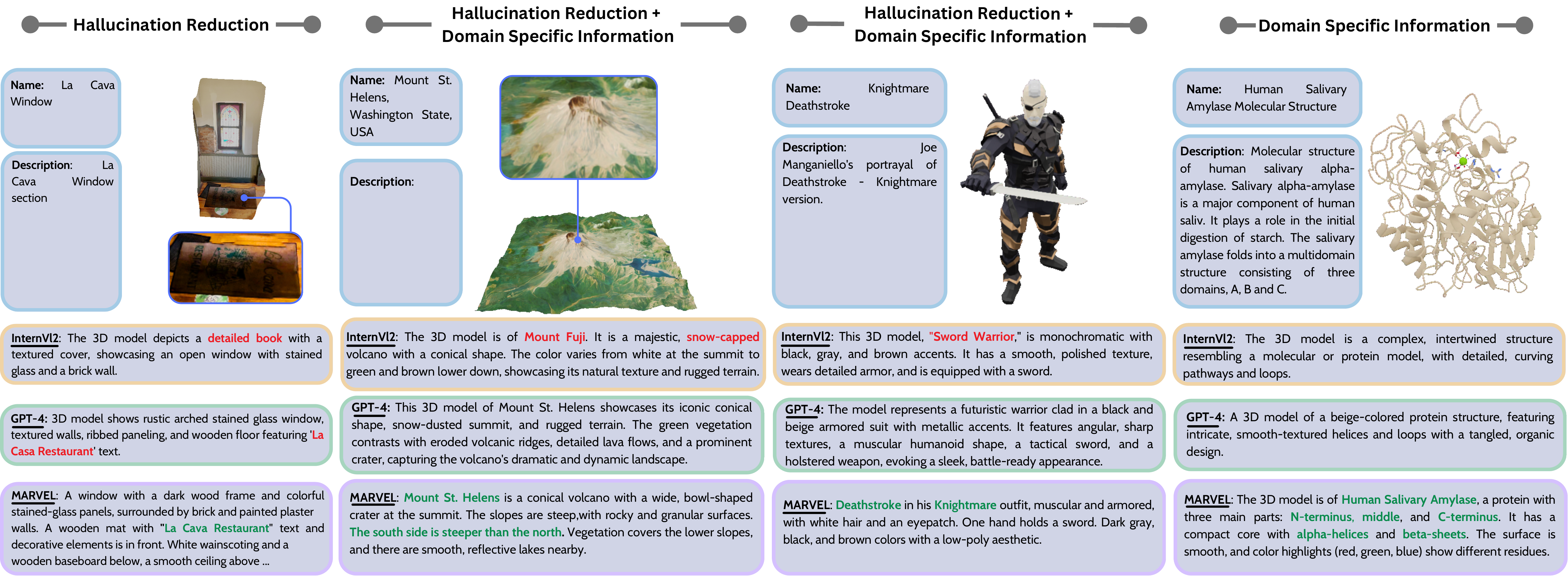}
    \caption{Effect of including human metadata, highlighting improvements in descriptive accuracy and contextual relevance compared to outputs generated without metadata, even when using state-of-the-art models like GPT-4~\cite{achiam2023gpt} and InternVL2~\cite{internvl}. Metadata inclusion helps reduce hallucinations and enhances domain-specific understanding.}
    \label{fig:effect_metadata}
\end{figure*}

Figure~\ref{fig:effect_metadata} showcases examples where human-provided metadata from source datasets reduce VLM hallucination and enhances annotations with domain-specific information. To generate captions using InternVL2~\cite{internvl,internvl1.5} and GPT-4~\cite{achiam2023gpt}, we input the same multi-view images used for MARVEL annotations, instructing them to produce concise descriptions that include names, shapes, textures, colors, and contextual environments.

\noindent Examples $1$, $2$, and $3$ demonstrate how the inclusion of simple metadata (\eg \textit{\textquote{La Cava Window}}, \textit{\textquote{Mount St. Helens}}) significantly reduces VLM hallucination, resulting in more accurate captions. Example $4$ illustrates how metadata can support the generation of highly domain-specific information (\eg \textit{\textquote{alpha-helices and beta sheets}}, \textit{\textquote{N-terminus, middle, and C-terminus}}).

\begin{figure*}[h]
    \centering
    \includegraphics[width=1\linewidth]{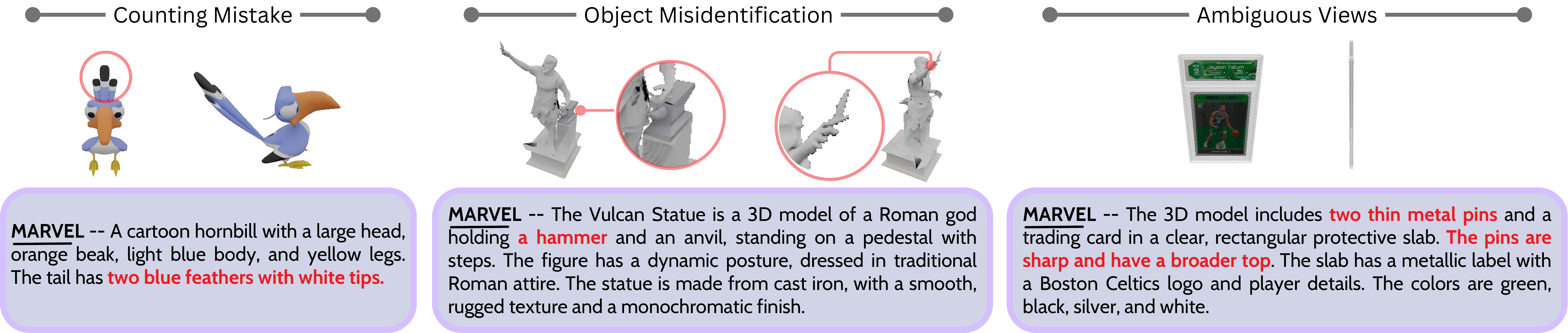}
    \caption{Failure cases of the MARVEL annotation pipeline. From left to right, the examples illustrate errors such as counting mistakes, object misidentification, and challenges with ambiguous views.}
    \label{fig:limitation}
\end{figure*}

\subsection{More 3D Captioning Results} 


We provide more qualitative comparisons of annotations, highlighting differences between the baseline~\cite{3dtopia,cap3d,ICML2024_LeveragingVLMs} and our proposed MARVEL-$40$M+ dataset. For consistency, we used only Level $4$ annotations, as their length closely matches that of the baselines. To improve clarity, we further categorized examples into distinct domains.

\begin{figure*}[h]
    \centering
    \includegraphics[width=0.9\linewidth]{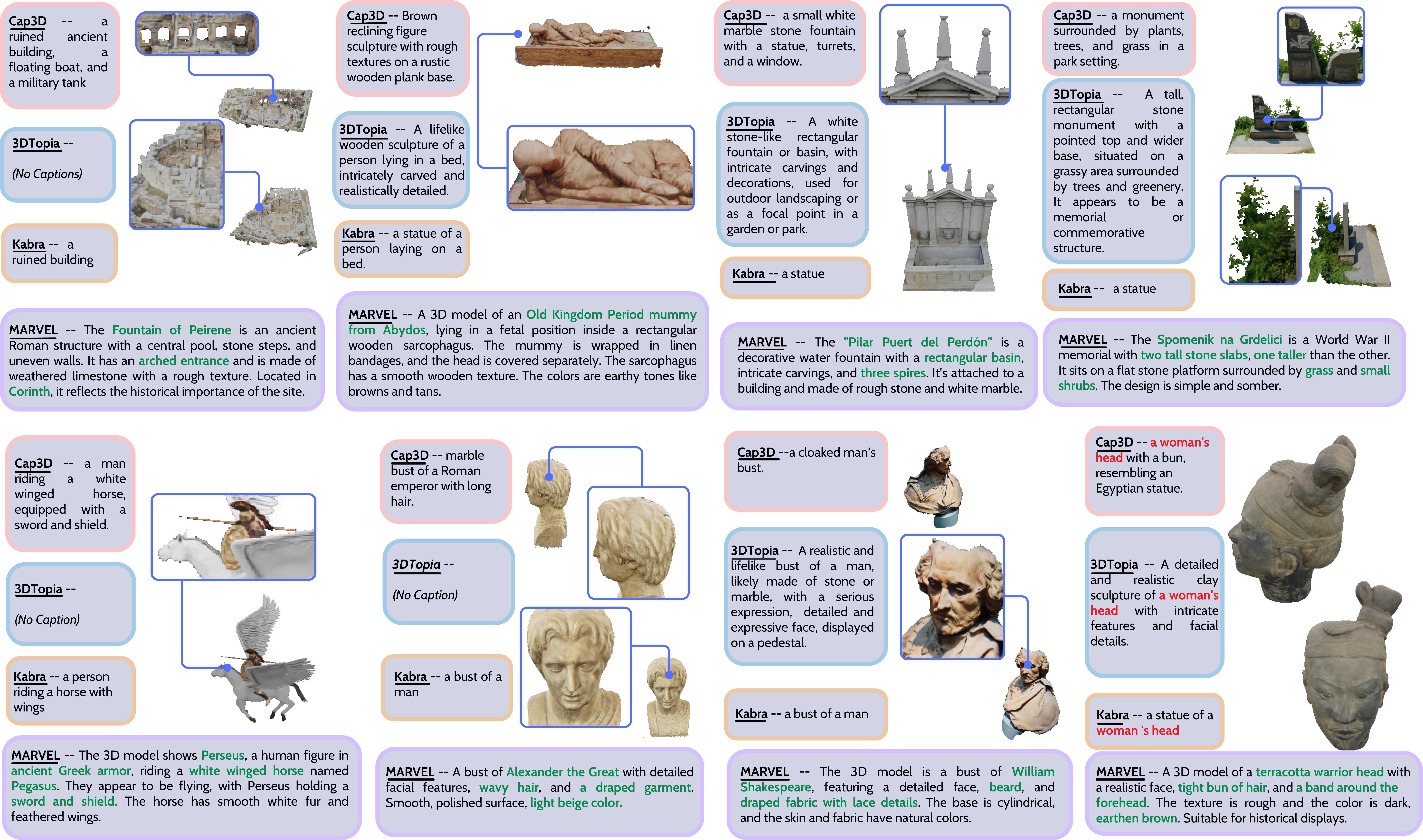}
    \caption{Qualitative comparison of 3D annotations across baselines~\cite{cap3d,3dtopia,ICML2024_LeveragingVLMs} and the proposed MARVEL-$40$M+ for \textit{historical elements including statues, places, memorials, etc}. Incorrect captions are highlighted in \textcolor{red}{\textbf{red}}, while important captions are highlighted in \textcolor{darkgreen}{\textbf{green}}.}
    \label{fig:obj1_history_comp}
\end{figure*}

\begin{figure*}[h]
    \centering
    \includegraphics[width=0.9\linewidth]{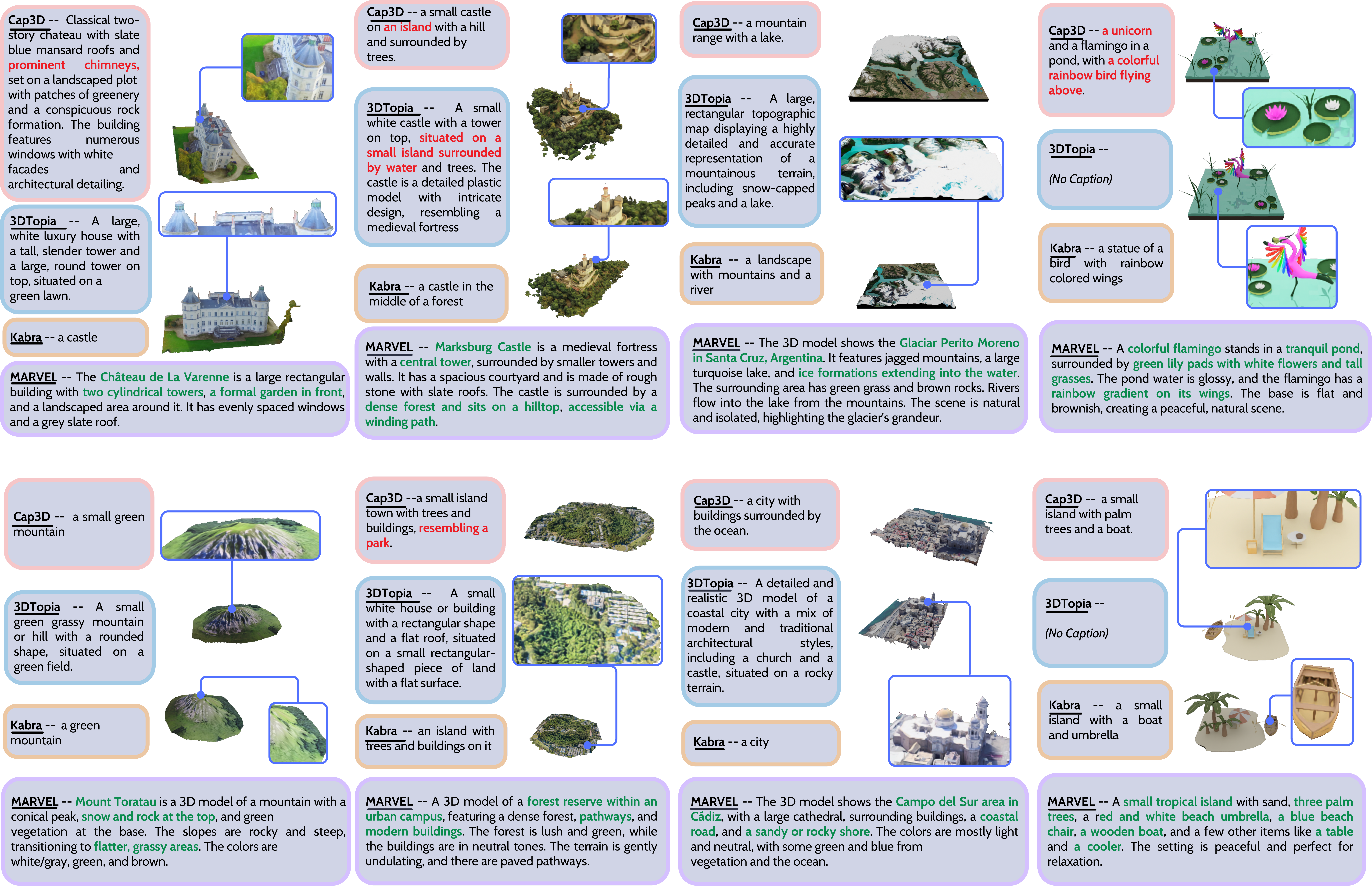}
    \caption{Qualitative comparison of 3D annotations across baselines~\cite{cap3d,3dtopia,ICML2024_LeveragingVLMs} and the proposed MARVEL-$40$M+ for \textit{diverse scenes including digital elevation maps, places, realistic or animated scenes}. Incorrect captions are in \textcolor{red}{\textbf{red}}, while important captions are in \textcolor{darkgreen}{\textbf{green}}.}
    \label{fig:obj1_scene_comp}
\end{figure*}

\begin{itemize}
    \item \textbf{Figure}~\ref{fig:obj1_machine_comp} showcases 3D models of automotive designs (e.g., \textit{cars, planes}) and CAD models.
    \item \textbf{Figure}~\ref{fig:obj1_character_comp} features iconic characters from \textit{anime, movies, and video games}.
    \item \textbf{Figure}~\ref{fig:obj1_animal_comp} illustrates biological elements such as \textit{animals, plants, and molecules}.
    \item \textbf{Figure}~\ref{fig:obj1_item_comp} includes diverse items ranging from \textit{everyday objects, essentials, food to luxury items}.
    \item \textbf{Figures}~\ref{fig:obj1_history_comp} and \ref{fig:obj1_scene_comp} depict historical artifacts (e.g., \textit{statues, memorials}) and various scenes (\eg \textit{digital elevation maps, realistic and animated scenes}) respectively.
\end{itemize}

\noindent As illustrated in the figures, MARVEL annotations offer more precise and domain-specific descriptions, leveraging accurate nomenclature and contextual terminology, surpassing the quality of the baseline datasets.

\subsection{More Multi-Level Examples}
\begin{figure*}[h]
    \centering
    \includegraphics[width=0.75\linewidth]{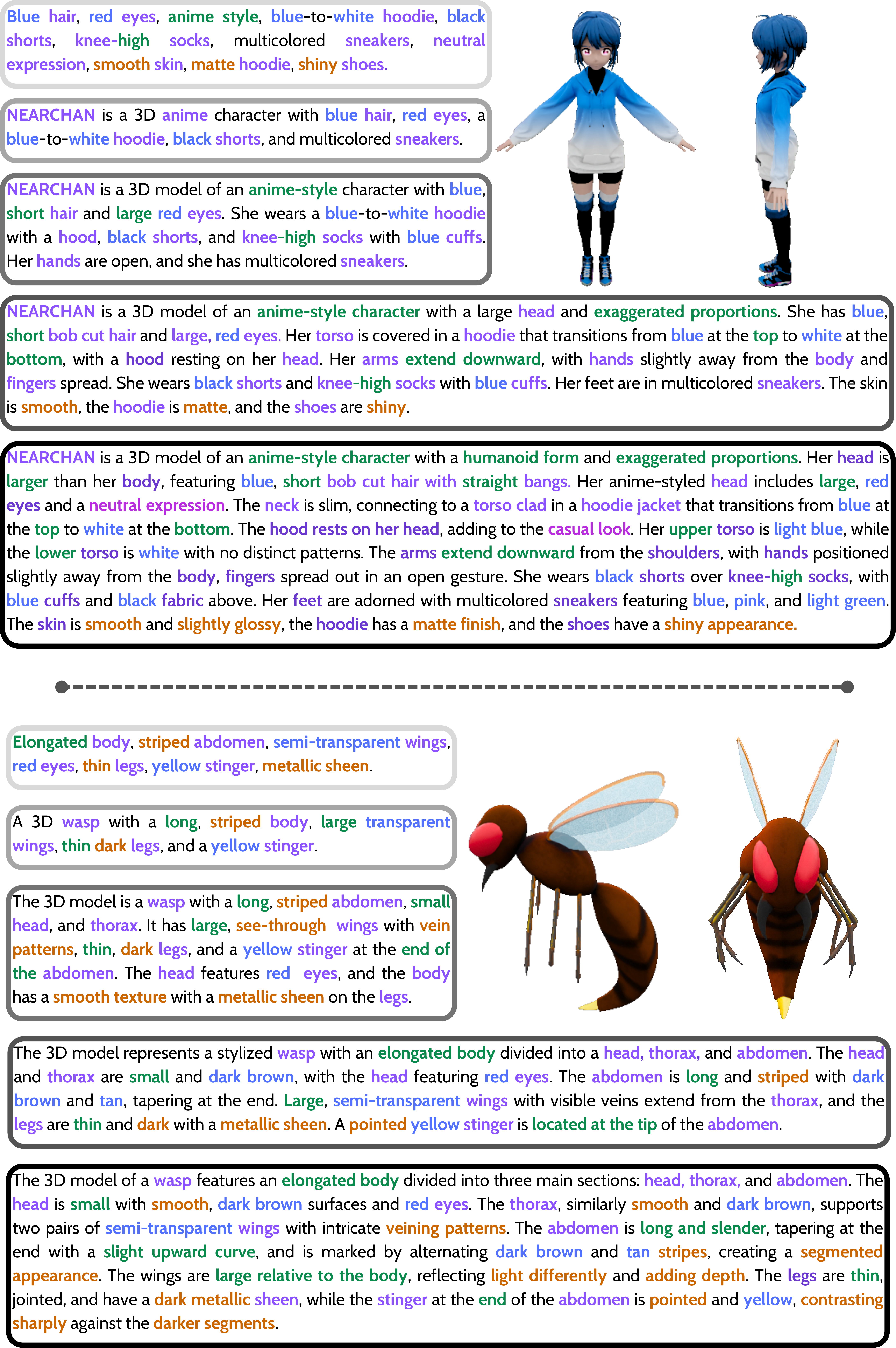}
    \caption{Multi-level annotation examples of MARVEL for the Objaverse~\cite{objaverse} dataset. Words corresponding to \textbf{\textcolor[HTML]{8c52ff}{Object and Components}} are highlighted in violet, \textbf{\textcolor[HTML]{0a8b4d}{Shape and Geometry}} in green, \textbf{\textcolor[HTML]{cc6600}{Texture and Materials}} in orange, \textbf{\textcolor[HTML]{5271ff}{Colors}} in blue, and \textbf{\textcolor[HTML]{b82edf}{Contextual Environment}} in purple. From top to bottom, we go from level-5 (Concise Tags) captions to level-1 (Comprehensive Description) captions.}
    \label{fig:objaverse_object_annotations}
\end{figure*}

\begin{figure*}[h]
    \centering
    \includegraphics[width=0.8\linewidth]{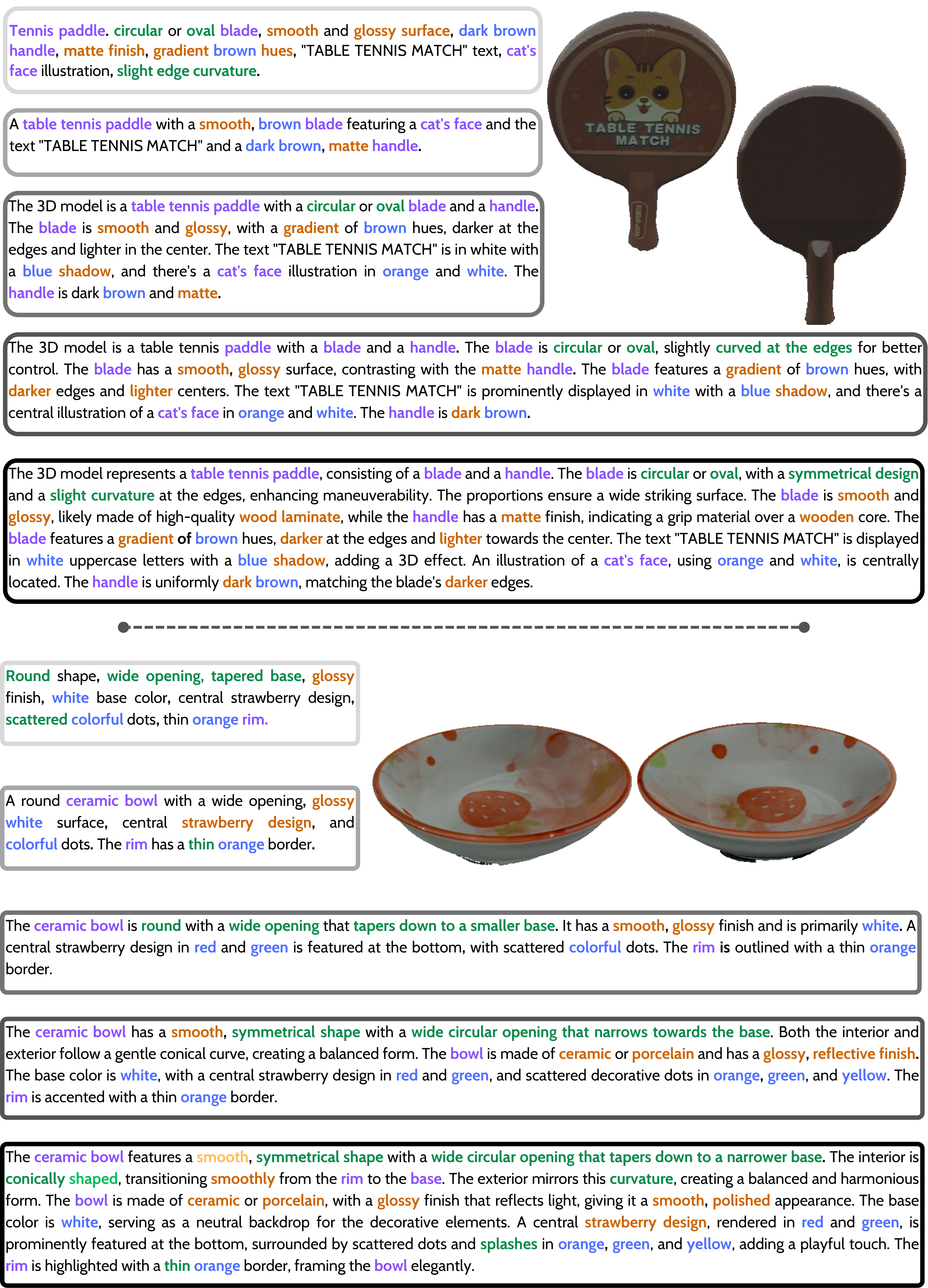}
    \caption{Multi-level annotation examples of MARVEL for the Omni-Object~\cite{omniobject3d} dataset. Words corresponding to \textbf{\textcolor[HTML]{8c52ff}{Object and Components}} are highlighted in violet, \textbf{\textcolor[HTML]{0a8b4d}{Shape and Geometry}} in green, \textbf{\textcolor[HTML]{cc6600}{Texture and Materials}} in orange, \textbf{\textcolor[HTML]{5271ff}{Colors}} in blue, and \textbf{\textcolor[HTML]{b82edf}{Contextual Environment}} in purple. From top to bottom, we go from level-5 (Concise Tags) captions to level-1 (Comprehensive Description) captions.}
    \label{fig:omni_object_annotations}
\end{figure*}

\begin{figure*}[h]
    \centering
    \includegraphics[width=0.8\linewidth]{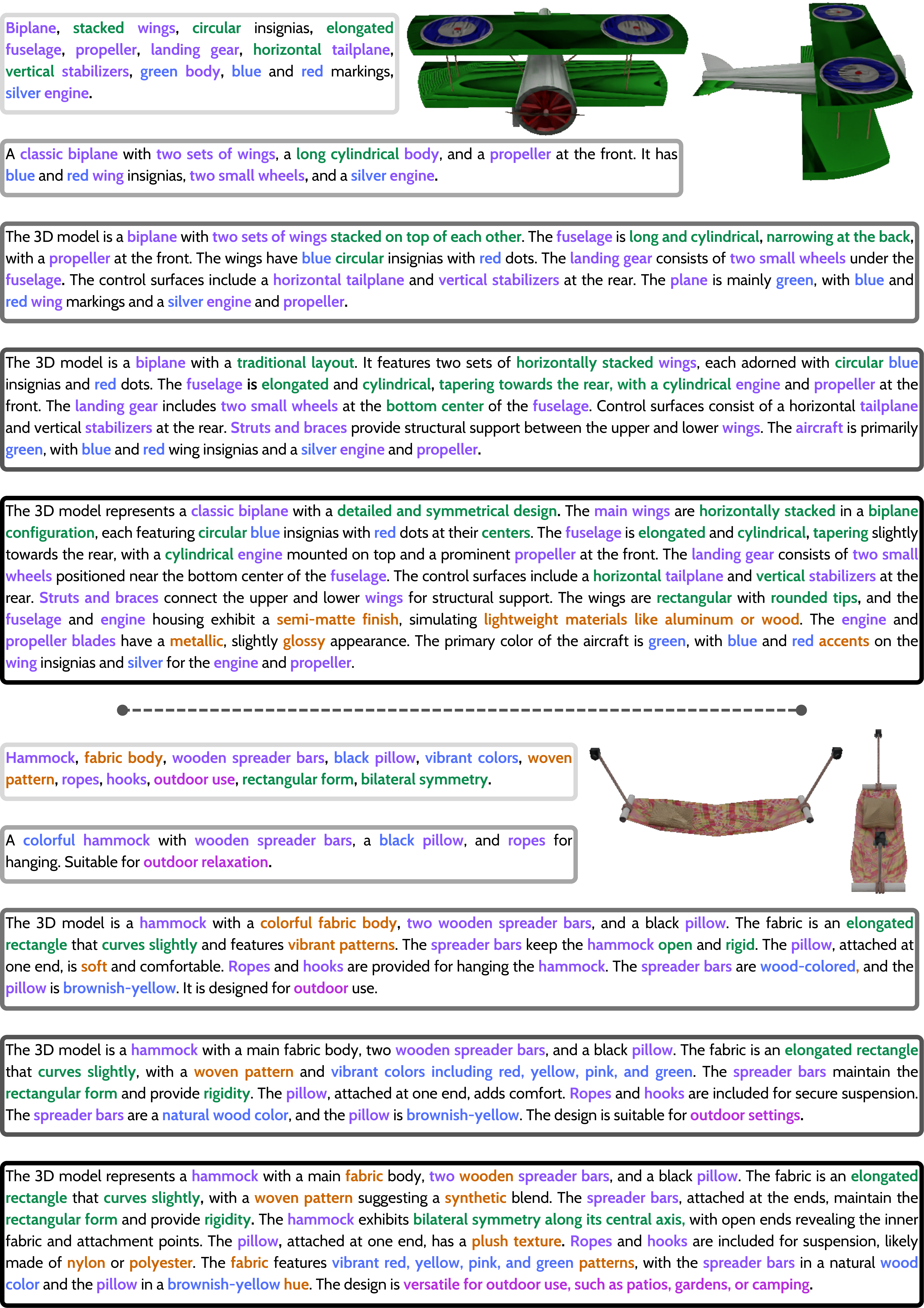}
    \caption{Multi-level annotation examples of MARVEL for the ShapeNet~\cite{shapenet} dataset. Words corresponding to \textbf{\textcolor[HTML]{8c52ff}{Object and Components}} are highlighted in violet, \textbf{\textcolor[HTML]{0a8b4d}{Shape and Geometry}} in green, \textbf{\textcolor[HTML]{cc6600}{Texture and Materials}} in orange, \textbf{\textcolor[HTML]{5271ff}{Colors}} in blue, and \textbf{\textcolor[HTML]{b82edf}{Contextual Environment}} in purple. From top to bottom, we go from level-5 (Concise Tags) captions to level-1 (Comprehensive Description) captions.}
    \label{fig:shape_net_annotations}
\end{figure*}

\begin{figure*}[h]
    \centering
    \includegraphics[width=0.8\linewidth]{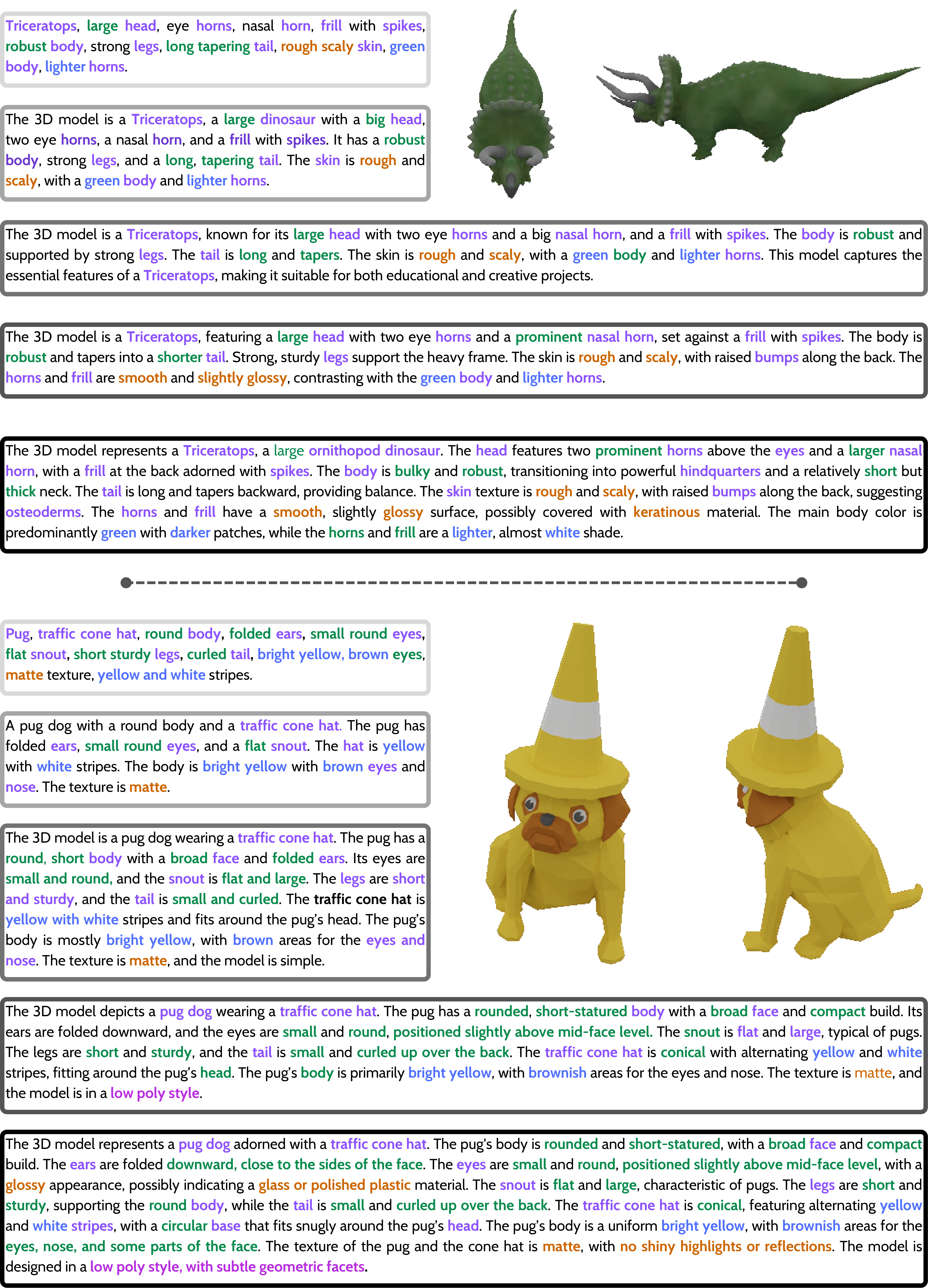}
    \caption{Multi-level annotation examples of MARVEL for the Toys4K dataset. Words corresponding to \textbf{\textcolor[HTML]{8c52ff}{Object and Components}} are highlighted in violet, \textbf{\textcolor[HTML]{0a8b4d}{Shape and Geometry}} in green, \textbf{\textcolor[HTML]{cc6600}{Texture and Materials}} in orange, \textbf{\textcolor[HTML]{5271ff}{Colors}} in blue, and \textbf{\textcolor[HTML]{b82edf}{Contextual Environment}} in purple. From top to bottom, we go from level-5 (Concise Tags) captions to level-1 (Comprehensive Description) captions.}
    \label{fig:toys_4k_annotations}
\end{figure*}

\begin{figure*}[h]
    \centering
    \includegraphics[width=0.80\linewidth]{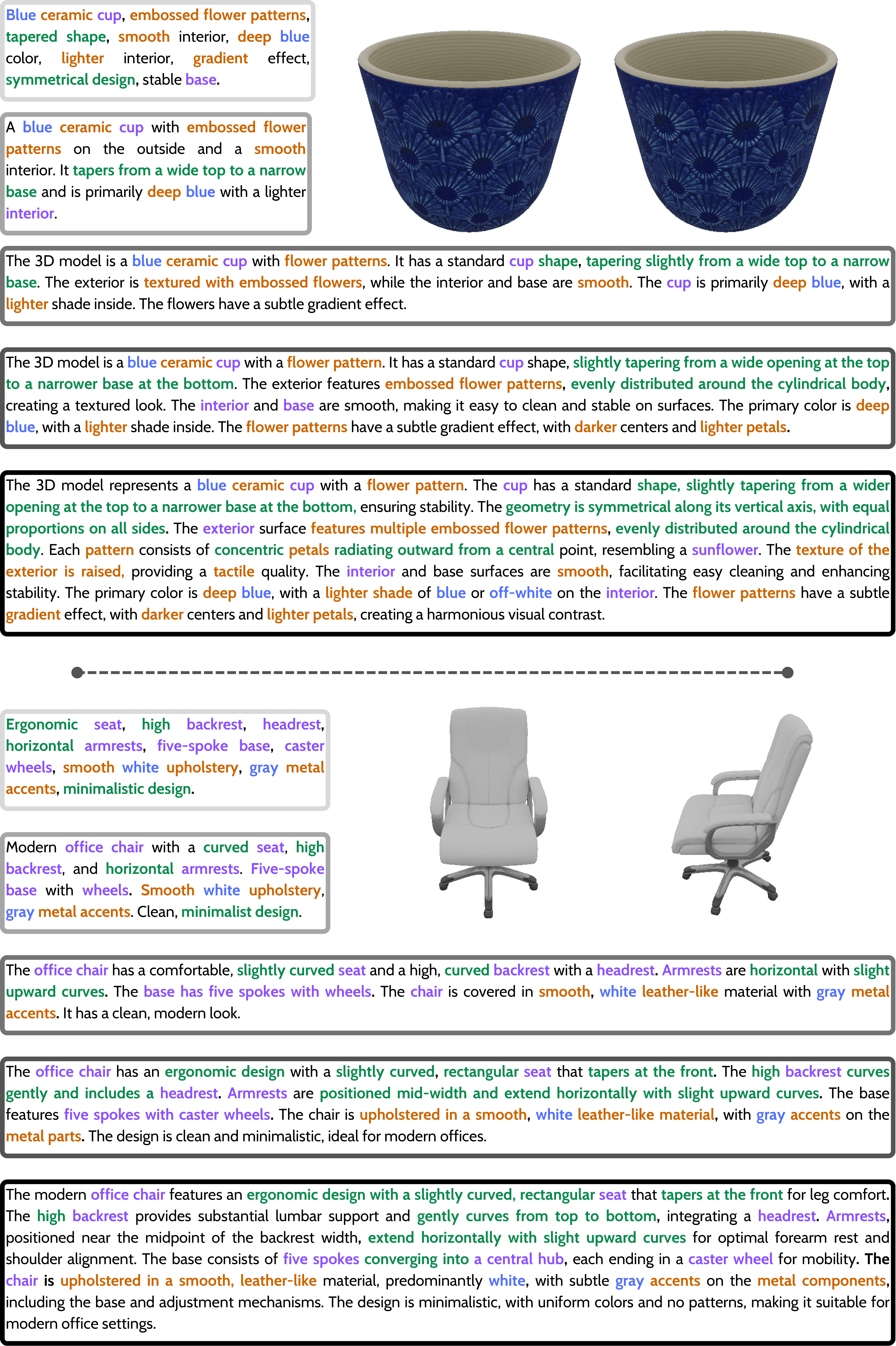}
    \caption{Multi-level annotation examples of MARVEL for the ABO (Amazon Berkeley Objects)~\cite{abo} dataset. Words corresponding to \textbf{\textcolor[HTML]{8c52ff}{Object and Components}} are highlighted in violet, \textbf{\textcolor[HTML]{0a8b4d}{Shape and Geometry}} in green, \textbf{\textcolor[HTML]{cc6600}{Texture and Materials}} in orange, \textbf{\textcolor[HTML]{5271ff}{Colors}} in blue, and \textbf{\textcolor[HTML]{b82edf}{Contextual Environment}} in purple. From top to bottom, we go from level-5 (Concise Tags) captions to level-1 (Comprehensive Description) captions.}
    \label{fig:abo_annotations}
\end{figure*}

\begin{figure*}[h]
    \centering
    \includegraphics[width=0.8\linewidth]{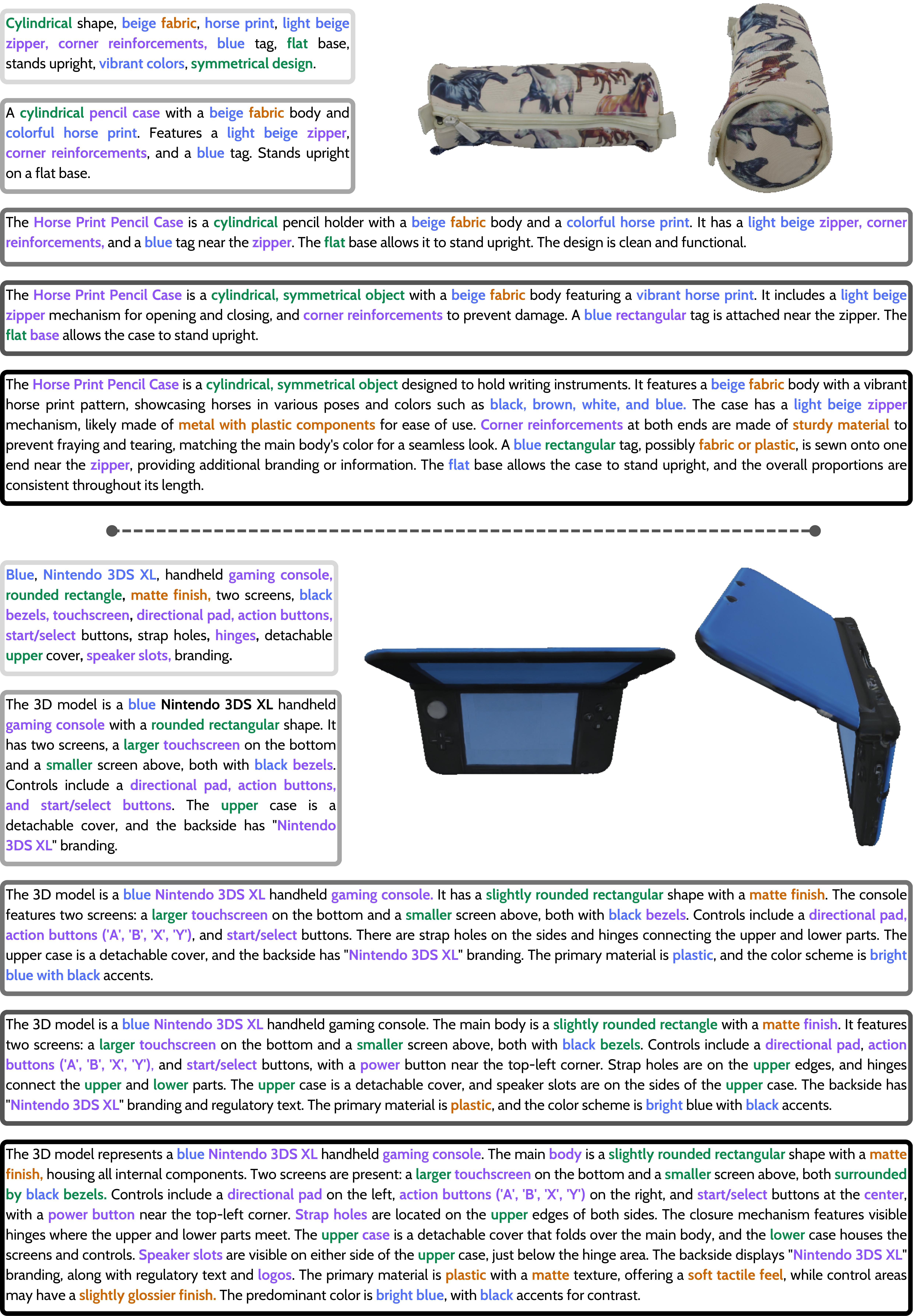}
    \caption{Multi-level annotation examples of MARVEL for the GSO (Google Scanned Objects)~\cite{gso} dataset. Words corresponding to \textbf{\textcolor[HTML]{8c52ff}{Object and Components}} are highlighted in violet, \textbf{\textcolor[HTML]{0a8b4d}{Shape and Geometry}} in green, \textbf{\textcolor[HTML]{cc6600}{Texture and Materials}} in orange, \textbf{\textcolor[HTML]{5271ff}{Colors}} in blue, and \textbf{\textcolor[HTML]{b82edf}{Contextual Environment}} in purple. From top to bottom, we go from level-5 (Concise Tags) captions to level-1 (Comprehensive Description) captions.}
    \label{fig:gso_annotations}
\end{figure*}

We present additional qualitative results showcasing our multi-level annotations across all seven datasets~\cite{objaverse,shapenet,pix3d,omniobject3d,toys4k,gso,abo}, with two examples per dataset - \textbf{Objaverse} (Figure~\ref{fig:objaverse_object_annotations}), \textbf{OmniObject3D} (Figure~\ref{fig:omni_object_annotations}),  \textbf{ShapeNet} (Figure~\ref{fig:shape_net_annotations}), \textbf{Toys4k} (Figure~\ref{fig:toys_4k_annotations}), \textbf{ABO} (Figure~\ref{fig:abo_annotations}) and \textbf{GSO} (Figure ~\ref{fig:gso_annotations}).

\subsection{More on Simple and Textureless Models}
\begin{figure*}[h]
    \centering
    \includegraphics[width=1\linewidth]{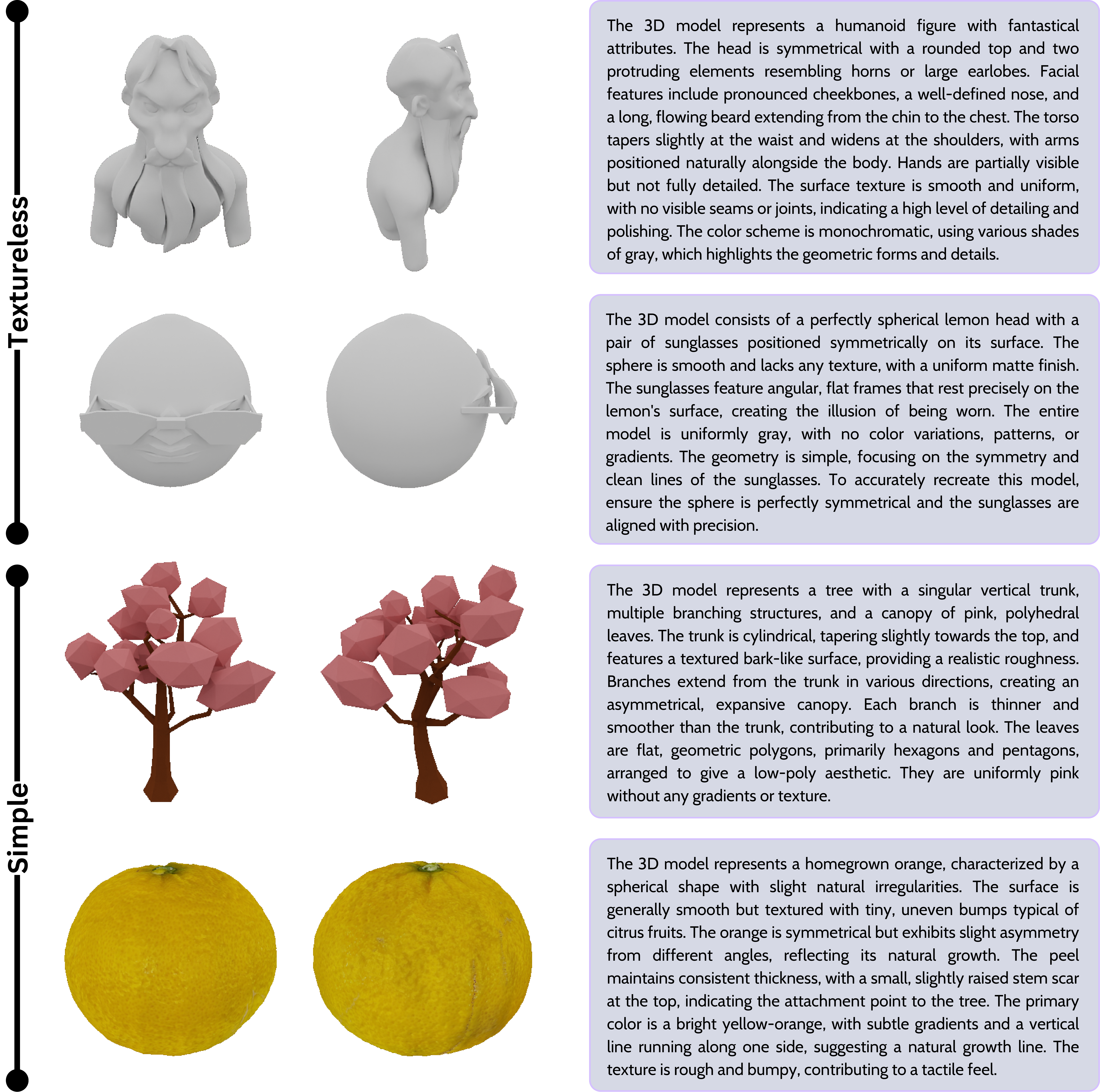}
    \caption{Examples illustrating MARVEL's robustness to simple and textureless models. Our annotation pipeline dynamically adjusts verbosity, ensuring concise yet accurate descriptions even when texture details are minimal or absent. (\textbf{Top}) Textureless models: a smooth humanoid figure with fantastical attributes, and a symmetrical, matte-finished spherical lemon head wearing sunglasses. (\textbf{Bottom}) Simple yet detailed models: a low-poly tree with geometric pink leaves, and a realistically textured homegrown orange showcasing subtle natural irregularities.}
    \label{fig:sim_and_tless}
\end{figure*}

Our annotation pipeline is robust and adaptable, effectively handling both simple and texture-less models by dynamically adjusting its descriptive verbosity. As illustrated in Figure \ref{fig:sim_and_tless}, the pipeline generates concise yet accurate annotations for texture-less models—such as the smooth, monochromatic humanoid figure and the symmetrical lemon head with sunglasses—by emphasizing their geometric precision and structural symmetry. Additionally, for simpler models like the low-poly tree with geometric leaves and the realistically textured orange, the pipeline adeptly captures essential details, highlighting subtle irregularities and primary shape characteristics. This flexibility ensures consistent annotation quality across diverse modeling scenarios.

\subsection{Need for Multi-Level-Structure}
Hierarchical structures are essential in AI. Examples include multi-resolution models in computer vision, such as feature pyramids, and hierarchical embeddings in NLP, such as document summarization. In 3D modeling, ShapeNet uses a hierarchical taxonomy \cite{shapenet} to improve adaptability across tasks. MARVEL similarly adopts a hierarchical design, ensuring task-specific granularity. By using a predefined hierarchy, MARVEL eliminates the need for repeated prompting. This reduces latency, computational costs, and inconsistencies associated with dynamically adjusting verbosity. Such dynamic adjustments would require multiple inference steps and additional processing, making them impractical for large-scale pipelines—even for future LLMs/VLMs. Additionally, dynamic generation introduces risks like semantic drift, information loss, and verbosity imbalance, decreasing annotation reliability. Our ablation study (Section 4.3B and Table 5) confirms that MARVEL’s structured verbosity effectively maintains essential details. It optimizes verbosity levels according to task requirements, as validated by cosine similarity and compression ratio.

\subsection{Failure Cases}

Figure~\ref{fig:limitation} presents examples of the failure cases discussed in Section 5 of the main paper, illustrating the challenges associated with using pretrained VLMs to generate dense descriptions of 3D models.

\subsection{More on MTLD Scores}
The \textbf{MTLD (Measure of Textual Lexical Diversity)} algorithm quantifies vocabulary diversity by segmenting a text whenever the \textit{Type-Token Ratio (TTR)}---the ratio of unique words to total words---drops below a fixed threshold (commonly 0.72). It processes the text both forwards and in reverse to reduce positional bias, and calculates the final MTLD score as the total number of words divided by the number of segments (called \textit{factors}). A low MTLD score indicates a repetitive vocabulary and low lexical diversity, while a high score reflects a rich and varied vocabulary. For instance, the repetitive string \texttt{"hello hello hello hello hello hello"} results in a low MTLD score of approximately 2.02, due to the lack of word variation. In contrast, the diverse sentence \texttt{"the quick brown fox jumps over the lazy dog"} yields a high MTLD score of around 22.68, as it contains many unique words. The pseudo-code for the algorithm is given in Algorithm \ref{alg:ago-1-MTLD} as seen in \cite{mtld}.
\vspace{0.3cm}
\begin{algorithm}[h]
\caption{MTLD Score \cite{mtld}}
\begin{algorithmic}[1]
\Function{MTLD}{text, min = 10}
    \State $forward \gets \Call{MTLDProcess}{text, min}$
    \State $reverse \gets \Call{MTLDProcess}{Reverse\_text, min}$
    \State \Return $(forward + reverse) / 2$
\EndFunction

\Function{MTLDProcess}{text, min}
    \State $factor \gets 0$
    \State $factor\_lengths \gets 0$
    \State $start \gets 0$
    \For{$x \gets 0$ \textbf{to} $length(text)-1$}
        \State $segment \gets text[start : x+1]$
        \If{$x+1 = length(text)$}
            \State $partial \gets \frac{1 - \Call{TTR}{segment}}{1 - 0.72}$
            \State $factor \gets factor + partial$
            \State $factor\_lengths \gets factor\_lengths + length(segment)$
        \ElsIf{$\Call{TTR}{segment} < 0.72$ \textbf{and} $length(segment) \ge min$}
            \State $factor \gets factor + 1$
            \State $factor\_lengths \gets factor\_lengths + length(segment)$
            \State $start \gets x+1$
        \EndIf
    \EndFor
    \State \Return $\frac{factor\_lengths}{factor}$
\EndFunction
\end{algorithmic}
\label{alg:ago-1-MTLD}
\end{algorithm}





\begin{figure*}[h]
    \centering
    \includegraphics[width=0.95\linewidth]{supp/assets/obj1_machine_comp.pdf}
    \caption{Qualitative comparison of 3D annotations across baselines~\cite{cap3d,3dtopia,ICML2024_LeveragingVLMs} and the proposed MARVEL-$40$M+ for \textit{automotive (cars, planes, etc) and CAD models}. MARVEL-40M+ provides more accurate and domain-specific annotations, compared to the baselines. Incorrect captions are highlighted in \textcolor{red}{red}, while important captions are highlighted in \textcolor{darkgreen}{green}.}
    \label{fig:obj1_machine_comp}
\end{figure*}

\begin{figure*}[h]
    \centering
    \includegraphics[width=1\linewidth]{supp/assets/obj1_character_comp.pdf}
    \caption{Qualitative comparison of 3D annotations across baselines~\cite{cap3d,3dtopia,ICML2024_LeveragingVLMs} and the proposed MARVEL-$40$M+ for \textit{popular anime, movie, and cartoon characters}. MARVEL-40M+ provides more accurate and domain-specific annotations, compared to the baselines. Incorrect captions are highlighted in \textcolor{red}{red}, while important captions are highlighted in \textcolor{darkgreen}{green}.}
    \label{fig:obj1_character_comp}
\end{figure*}

\begin{figure*}[h]
    \centering
    \includegraphics[width=1\linewidth]{supp/assets/obj1_animal_comp.pdf}
    \caption{Qualitative comparison of 3D annotations across baselines~\cite{cap3d,3dtopia,ICML2024_LeveragingVLMs} and the proposed MARVEL-$40$M+ for \textit{biological objects, including animals, plants, and molecular models}. MARVEL-40M+ provides more accurate and domain-specific annotations, compared to the baselines. Incorrect captions are highlighted in \textcolor{red}{\textbf{red}}, while important captions are highlighted in \textcolor{darkgreen}{\textbf{green}}.}
    \label{fig:obj1_animal_comp}
\end{figure*}

\begin{figure*}[h]
    \centering
    \includegraphics[width=0.97\linewidth]{supp/assets/obj1_item_comp.pdf}
    \caption{Qualitative comparison of 3D annotations across baselines~\cite{cap3d,3dtopia,ICML2024_LeveragingVLMs} and the proposed MARVEL-$40$M+ for \textit{diverse items including daily objects, essentials}. MARVEL-40M+ provides more accurate and domain-specific annotations, compared to the baselines. Incorrect captions are highlighted in \textcolor{red}{\textbf{red}}, while important captions are highlighted in \textcolor{darkgreen}{\textbf{green}}.}
    \label{fig:obj1_item_comp}
\end{figure*}

\begin{figure*}[h]
    \centering
    \includegraphics[width=0.85\linewidth]{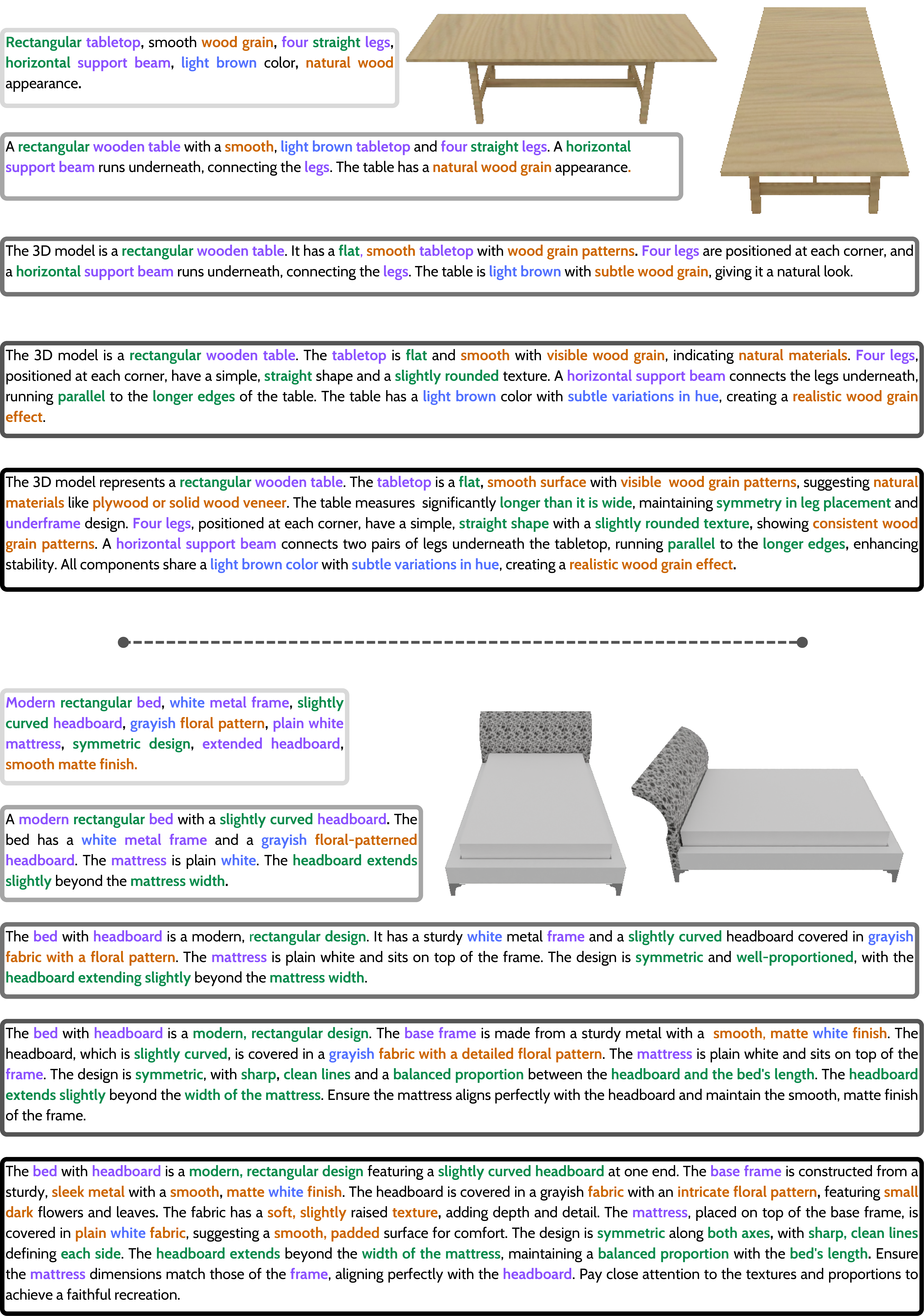}
    \caption{Multi-level annotation examples of MARVEL for the Pix3D~\cite{pix3d} dataset. Words corresponding to \textbf{\textcolor[HTML]{8c52ff}{Object and Components}} are highlighted in violet, \textbf{\textcolor[HTML]{0a8b4d}{Shape and Geometry}} in green, \textbf{\textcolor[HTML]{cc6600}{Texture and Materials}} in orange, \textbf{\textcolor[HTML]{5271ff}{Colors}} in blue, and \textbf{\textcolor[HTML]{b82edf}{Contextual Environment}} in purple. From top to bottom, we go from level-5 (Concise Tags) captions to level-1 (Comprehensive Description) captions.}
    \label{fig:pix_3d_annotations}
\end{figure*}

\begin{figure*}[h]
    \centering
    \includegraphics[width=1\linewidth]{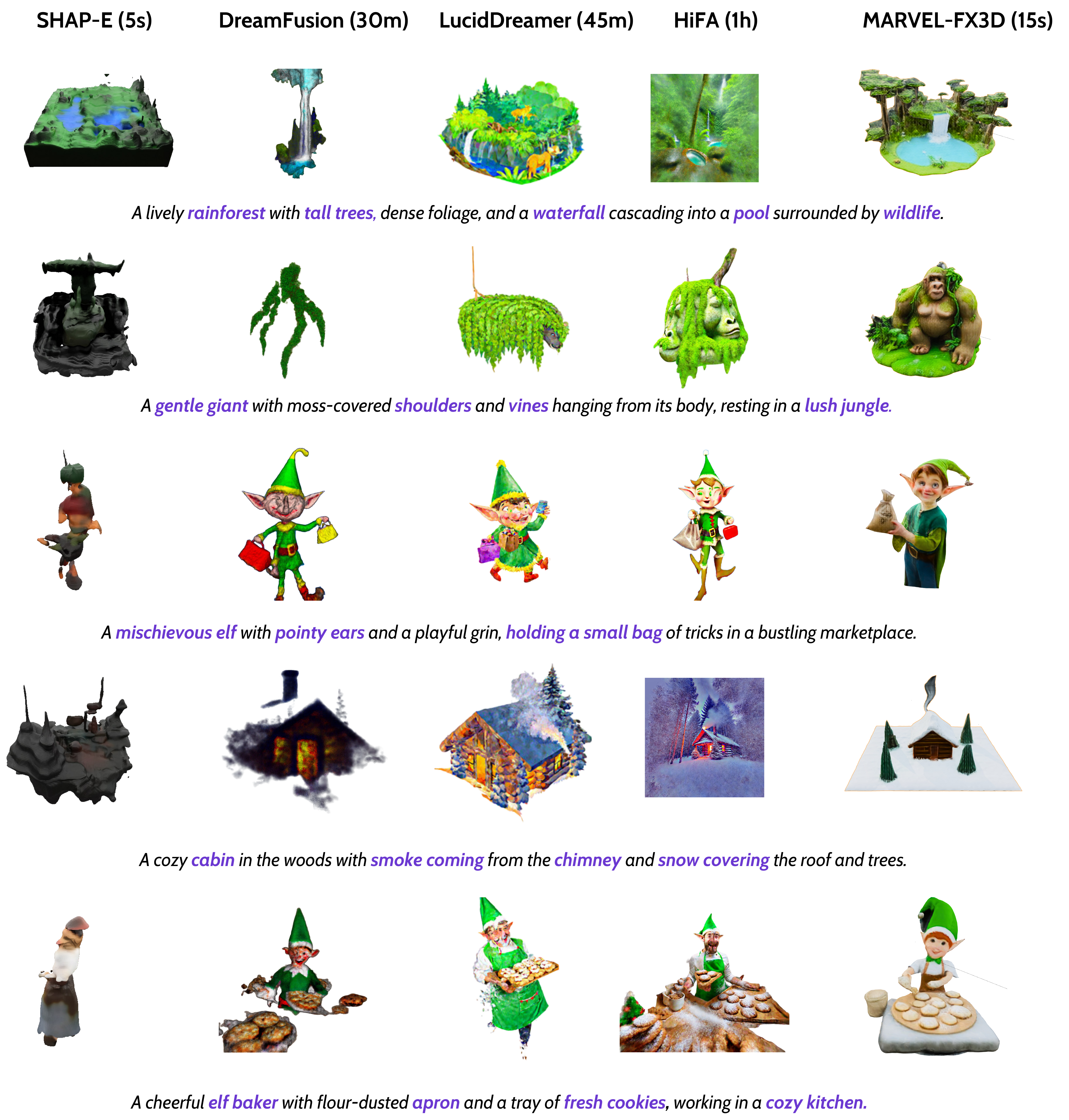}
    \caption{Qualitative Results for high fidelity TT3D generation on unseen prompts. From left to right, 3D models generated using Shap-E~\cite{shap_e}, DreamFusion~\cite{dreamfusion}, LucidDreamer~\cite{luciddreamer}, HiFA~\cite{hifa} and MARVEL-FX3D (ours).}
    \label{fig:baseline_1}
\end{figure*}

\begin{figure*}[h]
    \centering
    \includegraphics[width=1\linewidth]{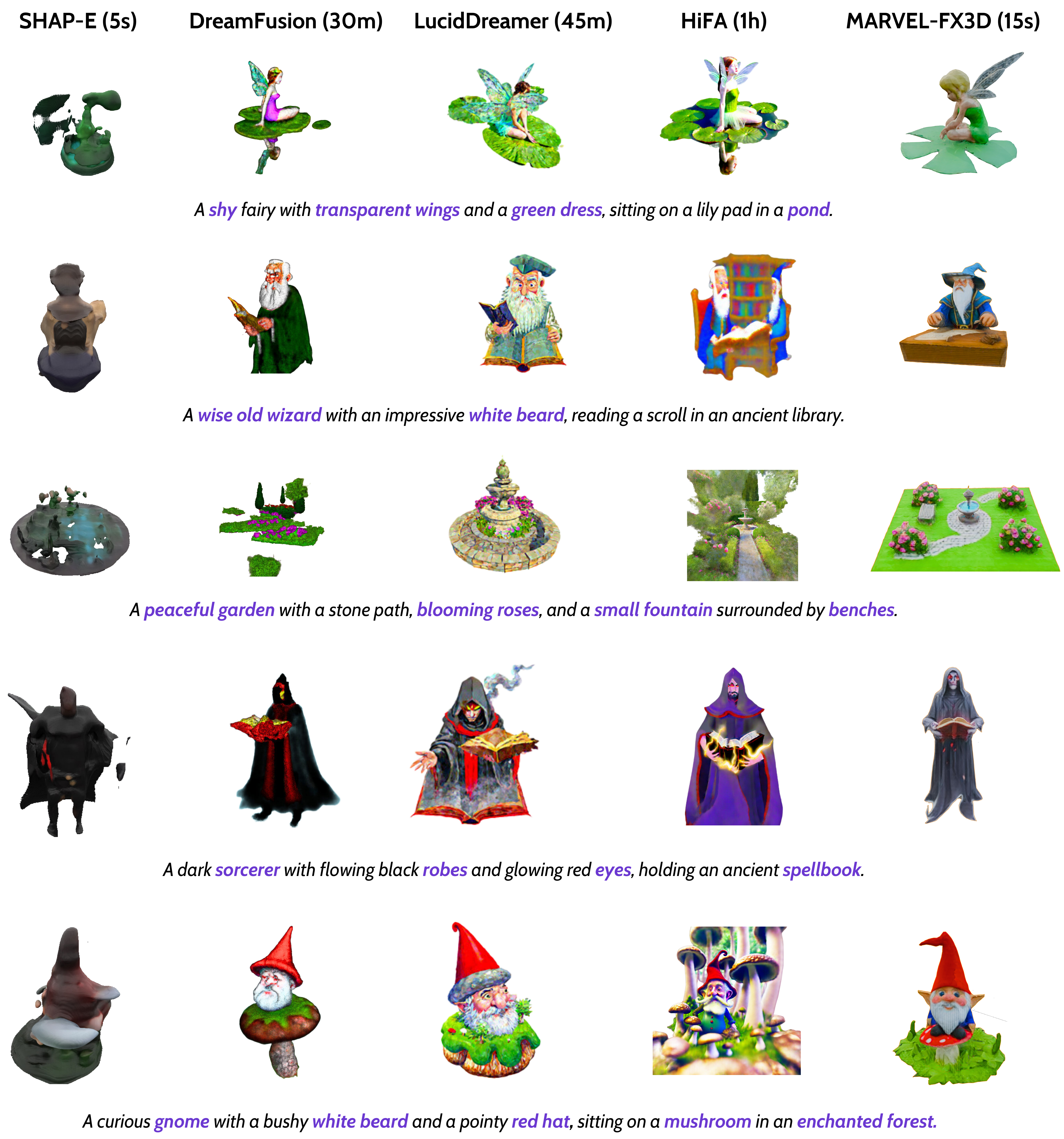}
    \caption{Qualitative Results for high fidelity TT3D generation on unseen prompts. From left to right, 3D models generated using Shap-E~\cite{shap_e}, DreamFusion~\cite{dreamfusion}, LucidDreamer~\cite{luciddreamer}, HiFA~\cite{hifa} and MARVEL-FX3D (ours).}
    \label{fig:baseline_2}
\end{figure*}

\section{Additional results of MARVEL-FX3D} \label{sec:tt3d}

\subsection{More Implementation Details}
As discussed in the main paper, MARVEL-FX3D is a two-stage pipeline. In the first stage, Stable Diffusion 3.5~\cite{sd3,sd3_huggingface} is fine-tuned. During each epoch, one annotation is sampled from five levels and paired with a randomly selected multi-view image for MSE loss calculation. During inference, CFG \cite{ho2022classifierfreediffusionguidance} is set to $7.5$, and 30 steps are used to balance speed and output diversity. 


\subsection{Baseline Adaptation}

We use the official implementations and pretrained models for Shap-E~\cite{shap_e} and Luciddreamer~\cite{luciddreamer}, training the latter for 3k steps. Dreamfusion~\cite{dreamfusion} and HIFA~\cite{hifa} are trained using the open-source threestudio~\cite{threestudio} implementation, with 10,000 and 24,000 steps, respectively, under default settings.

\subsection{More Text-to-3D Results}
\vspace{0.1cm}
Figures~\ref{fig:baseline_1} and~\ref{fig:baseline_2} showcase visual results of TT3D generation on unseen prompts. Using GPT-4~\cite{achiam2023gpt}, we generated 10 random prompts focused on shape and scene descriptions. As demonstrated, MARVEL-FX3D produces higher-fidelity 3D models from text prompts compared to the baseline methods.

\vspace{0.1cm}
\section{Discussion on Application of MARVEL}
\vspace{0.1cm}
The MARVEL-$40$M+ dataset, with its scale and diversity, serves as a powerful resource for text-to-3D tasks such as reconstruction, multi-view consistency, and compositional scene generation. A notable real-world use case, illustrated in Figure~\ref{fig:supp_teaser}, demonstrates how MARVEL-FX3D which is trained on MARVEL dataset enables rapid prototyping of diverse 3D objects from complex, fine-grained or simple text prompts. This capability facilitates the creation of intricate scenes, making it particularly valuable for applications in gaming, AR, and VR.


\end{document}